%% file: main.tex
\documentclass[10pt,twocolumn,letterpaper]{article}

\usepackage{wacv}              
\usepackage{cuted}
\usepackage{comment}
\usepackage{multirow}
\usepackage{float}
\usepackage[table]{xcolor}
\usepackage{booktabs}
\usepackage{threeparttable}
\definecolor{celgood}{RGB}{195,230,195}   
\definecolor{celbad} {RGB}{255,145,145}   
\definecolor{stdgray}{gray}{0.2}
\definecolor{cel@b}{RGB}{220,235,175}     
\definecolor{cel@c}{RGB}{250,235,160}     
\definecolor{cel@d}{RGB}{255,215,140}     
\definecolor{cel@e}{RGB}{255,190,155}     
\definecolor{cel@f}{RGB}{255,170,165}     
\newcommand{\gcell}[1]{%
  \ifdim #1 pt < 1.0pt \cellcolor{celgood}\else
  \ifdim #1 pt < 1.3pt \cellcolor{cel@b}\else
  \ifdim #1 pt < 1.6pt \cellcolor{cel@c}\else
  \ifdim #1 pt < 2.0pt \cellcolor{cel@d}\else
  \ifdim #1 pt < 2.4pt \cellcolor{cel@e}\else
  \ifdim #1 pt < 2.8pt \cellcolor{cel@f}\else
                        \cellcolor{celbad}%
  \fi\fi\fi\fi\fi\fi}
\newcommand{\gval}[1]{\gcell{#1}#1}
\newcommand{\gvalb}[1]{\gcell{#1}\textbf{#1}}
\newcommand{\gcello}[1]{%
  \ifdim #1 pt < 0.5pt \cellcolor{celgood}\else
  \ifdim #1 pt < 0.65pt \cellcolor{cel@b}\else
  \ifdim #1 pt < 0.80pt \cellcolor{cel@c}\else
  \ifdim #1 pt < 1.00pt \cellcolor{cel@d}\else
  \ifdim #1 pt < 1.20pt \cellcolor{cel@e}\else
  \ifdim #1 pt < 1.45pt \cellcolor{cel@f}\else
                         \cellcolor{celbad}%
  \fi\fi\fi\fi\fi\fi}
\newcommand{\gvalo}[1]{\gcello{#1}#1}
\newcommand{\gvalbo}[1]{\gcello{#1}\textbf{#1}}
\input{preamble}

\definecolor{wacvblue}{rgb}{0.21,0.49,0.74}
\usepackage[pagebackref,breaklinks,colorlinks,allcolors=wacvblue]{hyperref}

\title{FaceSnap: Real-Time Personalized Lightstage Facial Performance Capture}

\author{%
\begin{tabular}{ccc}
\textbf{Rukhshanda Hussain}\textsuperscript{1,2} &
\textbf{Noé Artru}\textsuperscript{1,2} &
\textbf{Emeline Got}\textsuperscript{2}\thanks{This work was done while the authors were at Ubisoft La Forge.} \\[0.3em]
\textbf{Luiz Gustavo Hafemann}\textsuperscript{2}\footnotemark[\value{footnote}] &
\textbf{Alexandre Messier}\textsuperscript{2} &
\textbf{Brandon Dearlove}\textsuperscript{2} \\[0.3em]
\textbf{Rafael M. O. Cruz}\textsuperscript{1} &
\textbf{Abdallah Dib}\textsuperscript{2} &
\textbf{Eric Granger}\textsuperscript{1} \\[0.8em]
\multicolumn{3}{c}{%
  \textsuperscript{1}\,École de Technologie Supérieure \quad
  \textsuperscript{2}\,Ubisoft La Forge%
}
\end{tabular}%
}
\newcommand{\ad}[1]{\textcolor{red}{#1}}

\begin{document}
\maketitle

\begin{strip}
    \centering
    \vspace{-22pt}
    \includegraphics[width=1\textwidth]{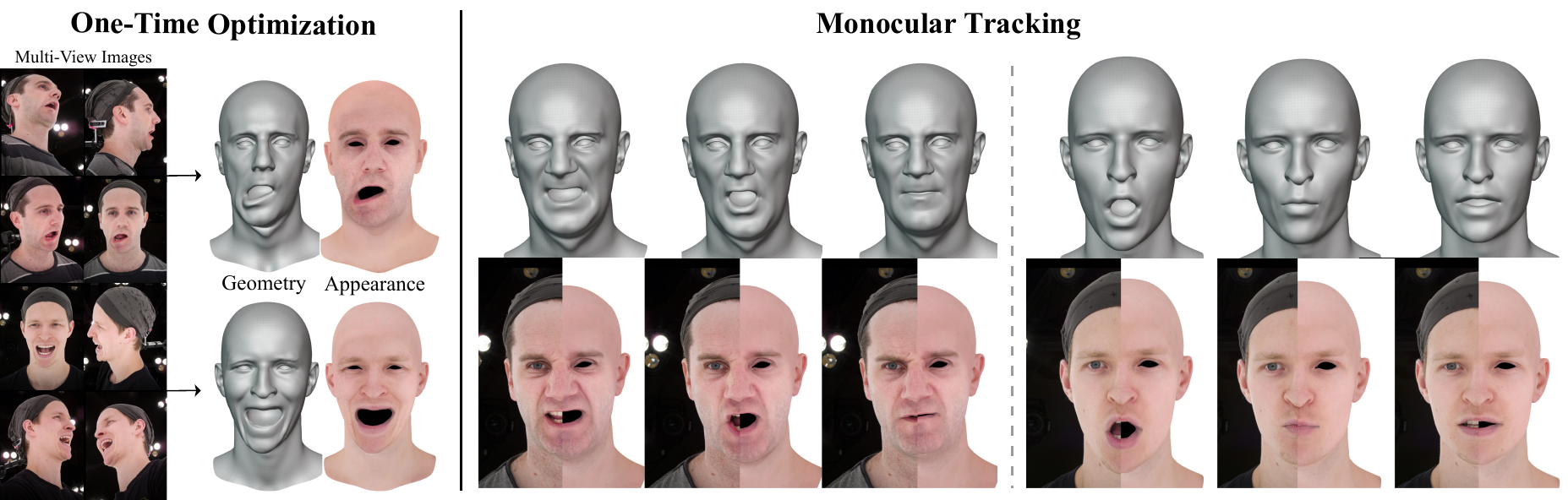}
    \captionof{figure}{FaceSnap produces high-fidelity geometry and appearance with topologically consistent results. \textbf{Left:} A one-time multi-view optimization builds a personalized subject model. \textbf{Right:} The personalized model enables real-time facial performance capture of new sequences from a single camera.
    \label{fig:teaser}}
\end{strip}

\begin{abstract}
Lightstage facial capture produces production-quality digital humans, but it is resource and labor-intensive. Multi-camera setups, hours of computation, and massive data storage create bottlenecks that hinder iterative workflows. This paper introduces FaceSnap, an end-to-end framework that streamlines capture via a two-stage approach. First, a one-time multi-view optimization from a range-of-motion sequence builds a personalized model encoding both geometry and expression-dependent appearance. This model then enables high-fidelity real-time facial performance capture from a single monocular lightstage camera, with no further multi-view capture required. FaceSnap jointly estimates geometry and dynamic 4K texture at 83 fps. The 4K texture is produced by a novel personalized residual upscaler that recovers subject-specific high-frequency detail, which generic upscalers fail to capture.
FaceSnap achieves geometric accuracy competitive with full per-frame multi-view optimization while outperforming feed-forward methods trained on production-quality 3D data, all from a single camera view. Finally, we introduce Multi4D, a public benchmark for evaluating 4D facial reconstruction methods in lightstage environments, enabling topology-invariant geometric comparison across methods.
\end{abstract}
\vspace{-14 pt}
\section{Introduction}
\label{sec:intro}

Modern game engines and high‑end visual effects pipelines increasingly demand digital human characters of unprecedented realism. To achieve this, lightstages \cite{debevec2012light}, equipped with dozens of synchronized cameras and controllable illumination, have become the standard for capturing facial geometry and reflectance. While lightstages capture highly detailed data, the associated workflows remain resource-intensive and time-consuming:  camera and light calibration require several hours, photogrammetric processing of hundreds of high-resolution images calls for days of computation and terabytes of storage, and manual post-processing is still required to register, clean, and retopologize meshes into production‑ready assets. Prior efforts to address these challenges fall into two main categories: generalized and subject-specific approaches. Generalized methods aim to work across individuals without per-subject optimization. This includes both lightstage-trained models, such as \cite{liu2022rapid, li2021topologically, Bolkart2023Tempeh}, which depend on extensive lightstage datasets and incur high capture costs, and "in-the-wild" approaches \cite{danvevcek2022emoca,feng2021learning, deng2019accurate, wang20243d, retsinas20243d, thies2016face2face, ploumpis2020towards, gecer2019ganfit, bas2017fitting, chang2018expnet, Zhang_2023_ICCV} that leverage publicly available monocular images to avoid specialized hardware. The former requires extensive capture sessions in the lightstage, while the latter simplifies capture, but still fails to reproduce the fine geometrical details and accurate reflectance needed for production-quality results.

In contrast, subject-specific methods \cite{yuan2024gavatar,giebenhain2024npga,saito2024relightable,qian2024gaussianavatars,teotia2024gaussianheads} approach facial reconstruction by optimizing for individual subjects, producing high-fidelity results without requiring extensive multi-subject datasets. 
Among these, Gaussian splatting-based methods \cite{kerbl20233d} provide visually impressive renderings but produce assets incompatible with mesh-based production pipelines. While some approaches attempt to extract registered, engine-friendly meshes from these splatting volumes \cite{li2024topo4d}, they remain offline procedures that require the full lightstage for every capture, along with per-session optimization. As a result, a practical bottleneck persists: capturing each new facial performance requires repeating the full multi-view capture and processing pipeline. Some methods reduce this repeated cost by learning a personalized, video-drivable asset \cite{zhang2022video}, but rely on synchronized high-speed photometric capture, a specialized configuration that limits adoption. 
This paper focuses on building a more standard multi-camera capture setup and introduces several practical simplifications to facilitate this style of capture.

We introduce FaceSnap, an end-to-end framework that amortizes the cost of high-fidelity facial capture. It involves a single, one-time multi-view session in the lightstage, which builds a personalized tracking model, after which every subsequent performance is captured in real time from a single lightstage camera. By front-loading the expensive multi-camera step into a reusable model, FaceSnap eliminates the need for repeated full-array captures that make current workflows impractical for iterative production. \autoref{fig:teaser} shows results for both stages.

\textbf{The key contributions of our work are:}
\textbf{(1)} FaceSnap, a fully automated, end-to-end pipeline for facial performance capture, in which a one-time multi-view optimization builds a personalized tracking model that enables real-time tracking from a single monocular lightstage camera. This amortizes the cost of multi-camera capture into a reusable model, eliminating the need for repeated full-array sessions that make current workflows impractical for iterative production.
\textbf{(2)} A novel one-time optimization that, through our regularization scheme, outperforms state-of-the-art reconstruction methods that require multi-array cameras.
\textbf{(3)} A real-time tracker, driven by the personalized model, that estimates geometry and 4K dynamic appearance from an expression code alone. At its core, a personalized two-stage residual upscaler recovers subject-specific high-frequency detail, allowing the tracker to outperform monocular baselines in texture fidelity while running in real time.
\textbf{(4)} Multi4D, a public benchmark based on Multiface \cite{wuu2022multiface} for 4D facial tracking in lightstage environments that enables topology-agnostic comparison across methods. This gap has prevented rigorous evaluation in prior work. Multi4D provides raw scan sequences, evaluation metrics source code for reproducible research.

\section{Related Works}
\begin{figure*}[t]
    \centering
    \includegraphics[width=0.9\linewidth]{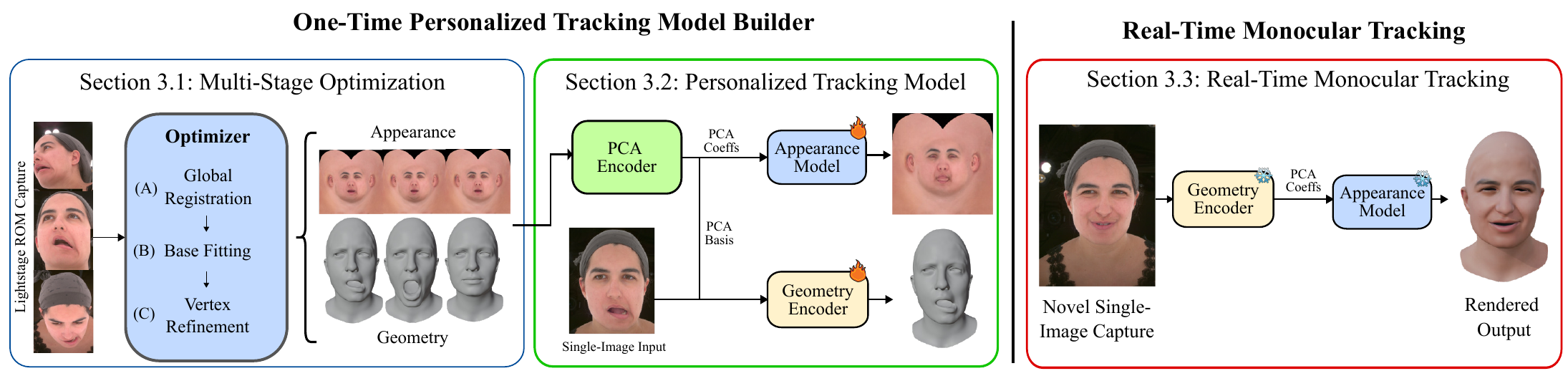}
    \caption{Overview of the FaceSnap method. From multi-view Lightstage captures, FaceSnap: (i) performs multi-stage optimization to obtain registered facial animation, and (ii) trains subject-specific models that estimate geometry and texture. These models enable high-fidelity monocular face capture of new performances.
    }
    \label{overall}

\end{figure*}
While lightstage capture provides high-fidelity facial data, 
obtaining meshes in a pre-defined topology
remains challenging. Production workflows require manual mesh retopology using tools like Wrap4D \cite{faceform2025}. Conventional optimization-based approaches \cite{cao2013facewarehouse, booth20163d, ji2021light, hsieh2015unconstrained, bradley2010high, beeler2011high} automate this via multi-view stereo reconstruction and non-rigid ICP registration to deform canonical models \cite{o19993d}.

Recent implicit representations \cite{mildenhall2021nerf} have been explored for photorealistic head avatars \cite{lombardi2021mixture,gafni2021dynamic, chandran2024anatomically}, with Gaussian splatting methods \cite{yuan2024gavatar,giebenhain2024npga,saito2024relightable,qian2024gaussianavatars,teotia2024gaussianheads} showing impressive results. However, these lack explicit geometry and texture control for production pipelines. Topo4D \cite{li2024topo4d} addresses this by outputting engine-compatible meshes and textures, but requires re-optimization for each new sequence and multi-camera lightstage capture. Moreover, Topo4D struggles to track rapid expression changes.

\noindent\textbf{(a) Learning-based Methods:}
To address slow optimization-based methods, several data-driven approaches \cite{bolkart2023instant, li2021topologically,liu2022rapid, li2024grape, alp2017densereg,jung2021learning} have been proposed. 
These methods take multi-view images as input and directly decode meshes in a consistent topology.
However, these learning-based methods are limited by their dependence on large amounts of collected scans (e.g. 64 subjects in \cite{liu2022rapid,li2021topologically}), which are expensive to collect. Moreover, they are camera-sensitive and trained on specific lightstage configurations—when camera positions change, the model requires complete retraining on similarly extensive datasets \cite{li2024grape}, limiting their practical usability. 
To avoid the need for specialized hardware, monocular methods leverage large-scale face images available in-the-wild and parametric models like FLAME \cite{li2017learning} to reconstruct facial geometry from single images or videos \cite{feng2021learning, danvevcek2022emoca, wang20243d, retsinas20243d, thies2016face2face, ploumpis2020towards, gecer2019ganfit, bas2017fitting, chang2018expnet, Zhang_2023_ICCV, dib2021towards, Dib_2024_CVPR}. Methods like EMOCA \cite{danvevcek2022emoca} and DECA \cite{feng2021learning} predict 3DMM coefficients and use differentiable rendering \cite{laine2020modular, kato2020differentiable} to minimize a photometric loss. Some works go beyond  3DMM while still being regularized by it \cite{zhu2022beyond, wang20243d, dib2022s2f2}. While these approaches bypass lightstage capture requirements, the quality remains below production standards due to the limited representation of parametric morphable models.

\noindent\textbf{(b) Personalized Models for Performance Capture:}
Learning-based methods require expensive data, while optimization methods need re-processing for each capture session. Personalized models \cite{li2020dynamic, lombardi2018deep, zhang2022video, cao2021real, bharadwaj2023flare, baert2024spark} have been proposed to reduce capture sessions. As an example, SPARK \cite{baert2024spark} builds personalized models from in-the-wild videos without Lightstage data but achieves results below production quality. Using lightstage data, Lombardi et al. \cite{lombardi2018deep} proposed Deep Appearance Models from VR cameras, while Zhang et al. \cite{zhang2022video} produce high-quality meshes and textures with wrinkle maps. Our work differs in three ways: (1) methods from \cite{zhang2022video,lombardi2018deep} rely on pre-registered meshes from off-the-shelf trackers, whereas we provide end-to-end mesh tracking from lightstage data; (2) their focus is appearance modeling, while ours streamlines facial capture from lightstage for both geometry and textures; and (3) \cite{zhang2022video} uses different capture hardware than traditional lightstage, introducing additional hardware constraints.
\section{Proposed FaceSnap Method}
As illustrated in \autoref{overall}, FaceSnap's two-stage approach
first performs a one-time multi-view optimization on a range-of-motion
sequence. The registered meshes are used as priors to build a personalized subject
model encoding geometry and expression-dependent appearance
(Sec. ~\ref{ssec:optim}--\ref{ssec:subject_profile}), thus enabling
real-time tracking from a single Lightstage camera for any subsequent
performance (Sec. ~\ref{ssec:monocular_tracking}).

\subsection{One-time Multi-Stage Optimization}
\label{ssec:optim}

\begin{figure*}
    \centering
    \includegraphics[width=0.95\linewidth]{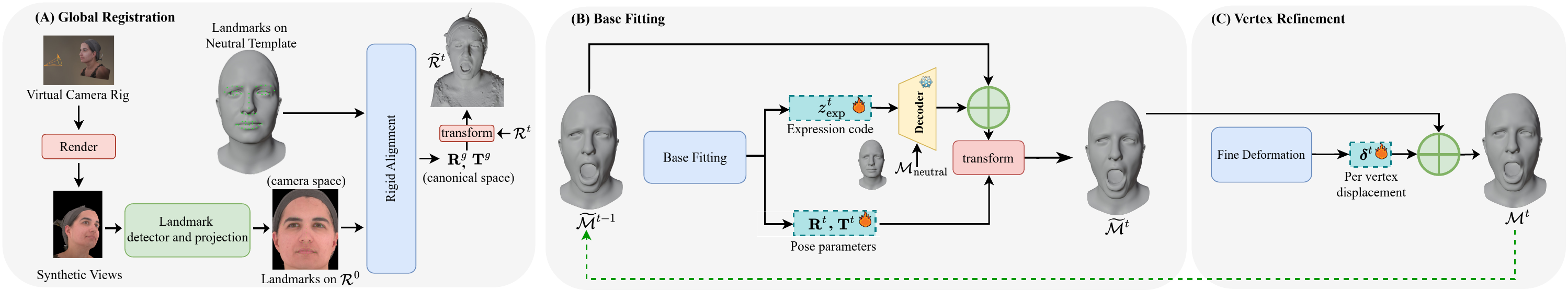}
    
    \caption{The multi-stage optimization of FaceSnap consists of (A) \emph{global registration}, which rigidly aligns the raw sequence to the canonical coordinate space, producing aligned sequence $\{\mathbf{\widetilde{R}}^t\}$ that serves as the target for subsequent stages. Then, for each frame, (B) \emph{base fitting} deforms the mesh $\mathcal{
    M}_\text{neutral}$ by optimizing rigid transformation and expression parameters constrained by a non-linear expression decoder. Finally, (C) \emph{vertex refinement} allows per-vertex displacements to obtain the registered mesh $\mathcal{M}^{t}$.}
    \label{tracking}
\end{figure*}

This step takes as input multi-view images $\{I^t_1, I^t_2, \ldots, I^t_N\}$
captured from $N$ cameras over time $t \in [0, S]$. A fully automated
photogrammetry step first produces a sequence of raw scans $\{\mathcal{R}^t\}$
with appearance textures $\{\mathcal{A}^t\}$, and a subject neutral mesh
$\mathcal{M}_{\mathrm{neutral}}$ in consistent topology is assumed provided. The goal is to generate registered meshes $\{\mathcal{M}^t\}$ in
canonical space with consistent topology while preserving photogrammetry
detail. Starting from $\mathcal{M}_{\mathrm{neutral}}$, our pipeline
comprises three stages: (1) global registration, (2) base fitting, and
(3) vertex refinement (see \autoref{tracking}). Building upon classical
mesh optimization~\cite{li2017learning}, our key improvements are a
non-linear expression model (Sec.~\ref{ssec:parametric_fitting}),
robust vertex refinement (Sec.~\ref{ssec:vertex_refinement}), and
specialized loss functions including a lip contour loss
(Sec.~\ref{ssec:loss}) that together enhance convergence stability
and geometric fidelity.

\noindent\textbf{Global Registration:} Procrustes
alignment~\cite{dryden2016statistical} is applied on neutral frame, $t=0$,
estimating a similarity transform $(\mathbf{R}_{\mathrm{g}},
\mathbf{T}_{\mathrm{g}}, s_{\mathrm{g}})$ that minimizes 3D landmark
distances between $\mathcal{R}^0$ and $\mathcal{M}_{\mathrm{neutral}}$,
then applying it to align the full sequence $\{\mathcal{R}^t\}$ to
canonical coordinates $\{\widetilde{\mathcal{R}}^t\}$
(\autoref{tracking}, Part A). Full details are provided in the
supplementary.

\subsubsection{Base Fitting}
\label{ssec:parametric_fitting}
Following global initialization, base fitting deforms the neutral mesh $\mathcal{M}_\mathrm{neutral}$ per frame by optimizing a low-dimensional latent expression code $\mathbf{z}^t_{\mathrm{exp}}$ (\autoref{tracking}\ad{, Part B}), which also regularizes the optimization. The deformation comes from a pretrained expression decoder $D_{\mathrm{exp}}$ (architecture follows \cite{josi2024serep}) that produces geometric deformations conditioned on the subject's neutral geometry, preserving identity-specific facial structure. The low-dimensional latent space provides effective regularization during optimization.
We also optimize a per-frame rigid transformation $(\mathbf{R}^t, \mathbf{T}^t) \in SE(3)$ to refine head pose and compensate for alignment imperfections. The deformed mesh at frame $t$ is:
\begin{equation}
\mathcal{\widetilde{M}}^t = \mathbf{R}^t \cdot \left[ \mathcal{M}_\mathrm{neutral} + D_{\mathrm{exp}}(\mathcal{M}_{\mathrm{neutral}}, z^t_{\mathrm{exp}}) \right] + \mathbf{T}^t
\label{eq:prior_mesh_deformation}
\end{equation}
We jointly optimize expression code \(z^t_{\mathrm{exp}}\) and pose parameters \((\mathbf{R}^t, \mathbf{T}^t)\) to minimize point-to-surface distance adaptively between transformed mesh \(\mathcal{\widetilde{M}}^t\) and rigidly aligned raw scan \(\mathcal{R}^t\), ensuring temporal consistency by initializing \(z^t_{\mathrm{exp}}\) and \((\mathbf{R}^t, \mathbf{T}^t)\) from the previous frame. Section \ref{ssec:loss} describes the complete loss function used for this stage.

\subsubsection{Vertex Refinement}
\label{ssec:vertex_refinement}
After the base fitting with the non-linear expression model, we proceed to a vertex refinement step to recover more subtle details and achieve better fitting (\autoref{tracking}\ad{, Part C}). While the parametric model provides a good foundation, it cannot accurately represent all possible expressions across different individuals. For this, we introduce per-vertex displacement vectors $\boldsymbol{\delta}^t \in \mathbb{R}^{N_v \times 3}$ that allow for more precise local adjustments at each frame $t$, where $N_v$ is the number of vertices. Each vertex $i$ has a 3D displacement vector $\delta^t_{i}$ optimized to capture fine-scale deformations. The final mesh is computed as:
\begin{equation}
\mathcal{M}^t =  \mathcal{\widetilde{M}}^{t} + \boldsymbol{\delta}^t
\label{eq:vertex_mesh_deformation}
\end{equation}
This unconstrained deformation can easily lead to implausible geometry and local minima. We therefore introduce regularization constraints (Section \ref{ssec:loss}), balancing detail capture and anatomical plausibility. The Part-based edge preservation, normal consistency, and Laplacian regularization prevent distortion while capturing subject-specific details beyond the parametric expression model.

\noindent\textbf{Automatic key framing with temporal smoothing. }
To prevent temporal drift, we employ automatic key framing that resets the per-vertex displacement field $\boldsymbol{\delta}^t$ when a neutral pose is detected. The current mesh $\mathcal{M}^t$ is compared against the neutral geometry $\mathcal{M}_{\mathrm{neutral}}$ using L2 distance on a frontal face mask. When the average distance falls below the threshold $\tau_f$, we trigger a reset and retroactively smooth displacements over the previous $N$ frames using linear decay $\alpha_i = \frac{i+1}{N}$ for frame $i$ back from the reset. This creates gradual transitions.

\subsubsection{Loss Functions} \label{ssec:loss}
The following loss functions are used in both optimization stages. For notation convenience, we denote $\mathcal{X}^t$ as the mesh being optimized: $\mathcal{X}^t = \mathcal{\widetilde{M}}^t$ during base fitting and $\mathcal{X}^t = \mathcal{M}^t$ during vertex refinement.\\
\textbf{Adaptive point-to-surface distance. }
We build on the point-to-surface distance objective of \cite{Bolkart2023Tempeh} and introduce two modifications for greater robustness and efficiency: 

(*1) Instead of using all mesh vertices, we uniformly sample N points $\mathcal{S}^t = \{\mathbf{s}_i\}_{i=1}^{N}$ from the raw scan. This sampling strategy ensures balanced spatial coverage regardless of the irregular vertex distribution in the raw scan, while significantly reducing computational cost. We also discard mouth and eye interior vertices by rendering the textured raw mesh and applying a face segmentation~\cite{zheng2022farl} to identify and discard vertices falling inside these regions, preventing these vertices from producing erroneous correspondences in the point-to-surface distance. (2) We introduce an adaptive correspondence weighting $w_i$ that automatically discards unreliable correspondences. The point-to-surface loss is then:
\begin{equation}
\mathcal{L}_{\text{p2s}} = \sum_{i=1}^{N} w_i \cdot \text{GM}\left(\|\mathbf{s}_i - \mathbf{q}_i\|^2, \sigma\right)
\end{equation}
where $\mathbf{q}_i$ is the closest point on $\mathcal{X}^t$ found by barycentric projection onto the nearest triangle face and $\text{GM}$ is the Geman–McClure robustifier \cite{geman1987statistical}. The adaptive weights are defined as: 

\begin{equation}
w_i =
\begin{cases}
1, & \text{if } \text{GM}\!\bigl(\|\mathbf{s}_i - \mathbf{q}_i\|^2, \sigma\bigr) < \tau_s, \\[6pt]
0, & \text{otherwise.}
\end{cases}
\end{equation}
with $\tau_s$ a distance threshold. This step removes corrupted or severely misaligned correspondences, particularly in regions where photogrammetry is unreliable (e.g., hair boundaries, specular highlights).

\noindent \textbf{Lip contour loss. }
Accurate lip tracking is critical for realistic facial animation. Traditional landmark-based approaches rely on a few sparse points that require fixed correspondences without semantic meaning, making them insufficient for capturing the complex deformations of lip contours. To address this limitation, we introduce a 2D chamfer distance loss that captures the complete outline of the lip without requiring any point correspondences.

Our method uses a facial segmentation network~\cite{khirodkar2024sapiens} to identify upper and lower lip regions, then applies contour detection to find the boundaries of each lip region. To extract the inner lip boundary, we group the detected contour points by x-coordinate and select the bottom-most points from the upper lip contour and the top-most points from the lower lip contour for each x-coordinate. This process extracts the inner edge of the lips, providing a  2D contour representation denoted as $Q$. The chamfer distance loss is formulated as:
\begin{equation}
\mathcal{L}_{\text{lip}} = \frac{1}{|P|}\sum_{p \in P} \min_{q \in Q} \|p - q\|^2 + \frac{1}{|Q|}\sum_{q \in Q} \min_{p \in P} \|q - p\|^2
\end{equation}
where $P$ represents the corresponding contour points from the registered mesh $\mathcal{X}^t$ back-projected to 2D using the camera perspective. This bidirectional formulation enables flexible contour matching without fixed correspondences, accurately capturing lip deformations.

\noindent \textbf{Regularization losses.}
To ensure robust and physically plausible mesh optimization, we employ three regularization terms. The \textit{Part-Based Edge Preservation Loss} $\mathcal{L}_{\text{edge}}$~\cite{Bolkart2023Tempeh} prevents sliding effects by constraining edge vectors between optimization iterations, with spatially-varying weights emphasizing critical facial regions like eyes and lips. The \textit{Laplacian Loss}  $\mathcal{L}_\text{lap}$~\cite{nealen2006laplacian} promotes smooth deformations by preserving local differential properties encoded in Laplacian coordinates, particularly important for challenging photogrammetry regions such as eyebrows and eyelids. The \textit{Normal Consistency Loss} $\mathcal{L}_\text{nc}$ encourages smooth surface reconstruction by penalizing inconsistent orientations between adjacent face normals. These regularization terms collectively maintain mesh topology consistency while allowing for natural deformations during the optimization. Full details are in the supp. material.\\
\textbf{Overall Objective Function}. The total loss function is formulated as a weighted combination of individual losses:
\begin{equation}
\label{eq:final_loss}
\mathcal{L}_{\mathrm{total}} = \lambda_1 \mathcal{L}_{\mathrm{p2s}} + \lambda_2 \mathcal{L}_{\mathrm{lip}} + \lambda_3 \mathcal{L}_{\mathrm{lap}} + \lambda_4 \mathcal{L}_{\mathrm{edge}} + \lambda_5 \mathcal{L}_{\mathrm{nc}}
\end{equation}
The weighting coefficients $\lambda_i$ are tuned differently for each optimization stage and provided in the supp. material.
\subsubsection{Texture Extraction}
To obtain high-quality appearance texture set $\{\mathcal{T}^t\}$, at 4K resolution,  for each registered mesh $\mathcal{M}^t$,
we use the spatial correspondence established during registration to map texture information from the raw scan $\mathcal{R}^t$ to the registered mesh $\mathcal{M}^t$. Specifically, we project the photogrammetry texture onto the consistent UV parameterization of the template topology. This is achieved by finding the nearest surface point on the raw scan for each vertex in the registered mesh and sampling the corresponding texture coordinates.
This results in a sequence of meshes with both geometric consistency and photorealistic textures.
\subsection{Personalized Tracking Model}
\label{ssec:subject_profile}

After optimization (\autoref{ssec:optim}) on the subject's ROM session, we obtain registered meshes $\{\mathcal{M}^t\}$ and textures
$\{\mathcal{T}^t\}$, used to train an expression dependent personalized tracker comprising a geometry model and two-stage appearance model.

\subsubsection{Geometry Model}\label{geom_model}
To create a compact geometric representation, we compress registered meshes $\{\mathcal{M}^t\}_{t=1}^{T}$ into a 256-dimensional PCA subspace. This representation retains over 99\% of geometric variance, indicating that facial expressions exhibit strong linear correlations that can be effectively captured without requiring more complex nonlinear models. Each frame is represented by coefficients $\mathbf{c}^t \in \mathbb{R}^{256}$. We tested 256 and 1024 dimensions; both retained $>$99\% variance with negligible quality difference, so we use 256 for efficiency. 

We train a ResNet encoder $E_\text{geom}: \mathbb{R}^{H \times W \times 3} \rightarrow \mathbb{R}^{256}$ that maps $512\times512$ monocular frames directly to PCA coefficients.To enable direct mesh reconstruction from these coefficients,  we append a frozen PCA decoder layer $F: \mathbb{R}^{256} \rightarrow \mathbb{R}^{3N_v}$ initialized with the PCA basis as weights and mean shape as bias. The encoder is trained exclusively on ROM sequences using the registered meshes from the optimization stage as ground truth with L2 loss: $\mathcal{L}_{track} = \|\mathcal{M}^t - F(E_\text{geom}(I^t))\|_2^2$, where $I^t$ is the input image.

\subsubsection{Dynamic Appearance}
\label{sec:appearance}

\begin{figure}[!t]
    \centering
    \includegraphics[width=\linewidth]{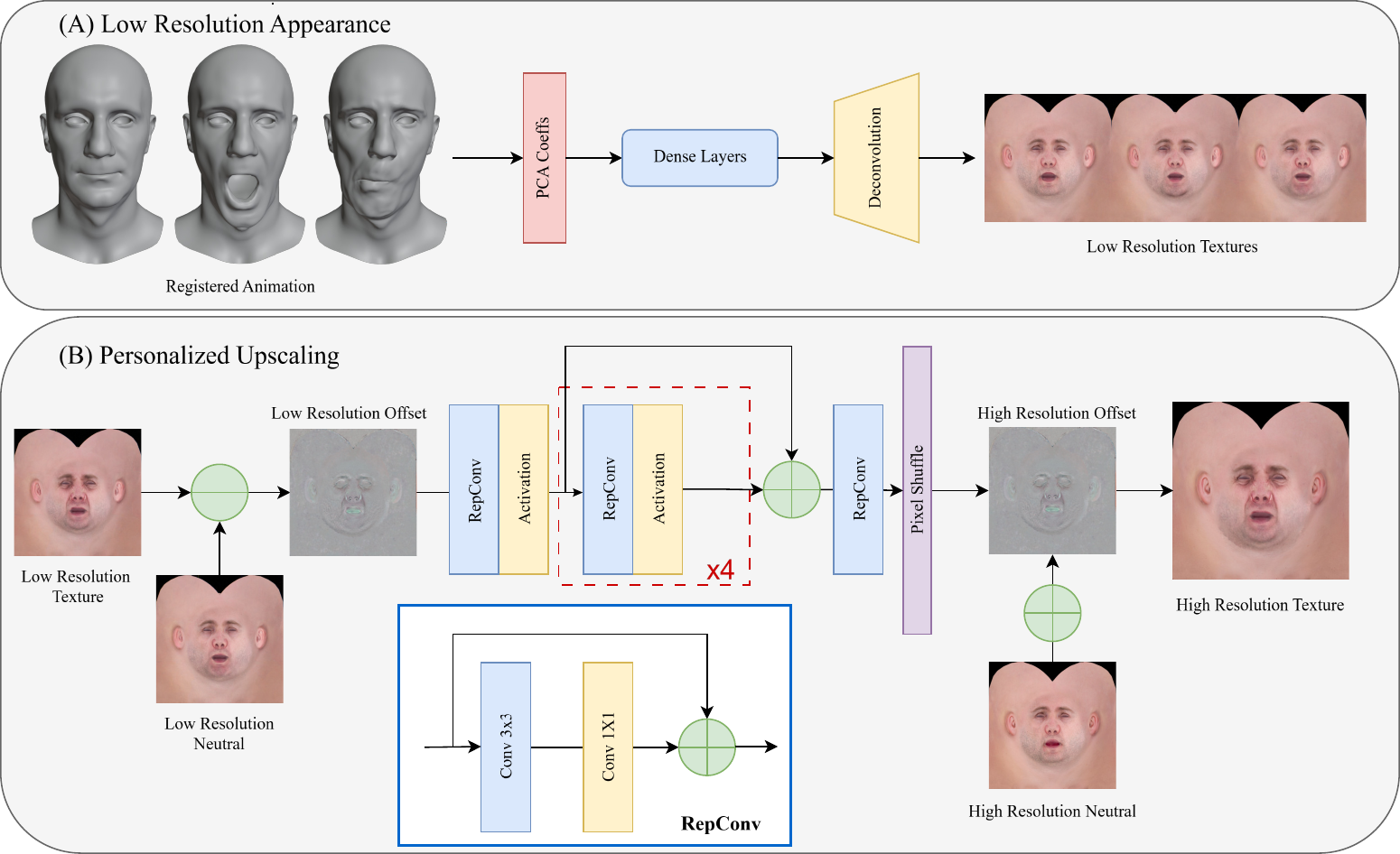}
    \caption{\textbf{Two-stage dynamic appearance pipeline.}
(A)~\textit{Low-res synthesis}: PCA coefficients $\mathbf{c}^t$ are decoded via deconvolution layers into a low-res dynamic texture $\hat{\mathcal{T}}^t$, supervised with masked L1 against ground-truth UV textures $\mathcal{T}^t$.
(B)~\textit{Personalized upscaling}: The offset $\hat{\mathcal{T}}^t - \mathcal{T}^{\text{LR}}_{\text{neut}}$ is lifted to 4K via residual RepConv blocks and added to $\mathcal{T}^{\text{HR}}_{\text{neut}}$, recovering fine-scale expression detail.
Both stages train solely on ROM data.}
            
            \label{app-full}
\end{figure}

We recover production-quality dynamic appearance in two stages: a lightweight expression-conditioned decoder that synthesizes low-resolution texture maps from the learned geometric representation, followed by a personalized upscaler that lifts these maps to 4K resolution in real time, as shown in \autoref{app-full}.
\vspace{-7pt}
\paragraph{Stage 1: Expression-Conditioned Texture Synthesis}
Given the subject profile built in \autoref{geom_model}, we train an expression-dependent texture decoder $D_{\text{tex}}: \mathbb{R}^{d} \rightarrow \mathbb{R}^{512 \times 512 \times 3}$ that generates texture maps conditioned solely on the geometric PCA coefficients $\mathbf{c}^t$ extracted from that profile. This design explicitly captures the correlation between facial geometry and appearance — wrinkles, specular highlights, and skin deformations are all encoded in $\mathbf{c}^t$ — without requiring image-based supervision at inference time. The decoder is trained with a masked L1 loss in UV space:
\begin{equation}
    \mathcal{L}_{\text{tex}} = \|\mathbf{M} \odot (\mathcal{T}^t - D_{\text{tex}}(\mathbf{c}^t))\|_1,
\end{equation}
where $\mathcal{T}^t$ is the ground-truth UV texture for frame $t$ and $\mathbf{M}$ is a frontal face mask that concentrates supervision on critical regions. No additional capture is required beyond the single Lightstage session used to build the subject profile.
\vspace{-10pt}
\paragraph{Stage 2: Personalized Residual Upscaler} \label{upscaler}
Off-the-shelf super-resolution models introduce smoothing artifacts on facial textures and fail to preserve high-frequency details such as skin pores and subsurface structures, as they have no knowledge of the subject's specific appearance. We therefore train a subject-specific upscaling network $U_{\theta}$ that lifts the 512$\times$512 output of Stage 1 directly to a 4096$\times$4096 texture.

Rather than predicting absolute pixel values, $U_{\theta}$ performs \textit{residual prediction}: it estimates a pixel-space offset $\boldsymbol{\Delta}^t$ over the subject's neutral high-resolution texture $\mathcal{T}^{\text{HR}}_{\text{neut}}$,
\begin{equation}
    \hat{\mathcal{T}}^t_{\text{HR}} = \mathcal{T}^{\text{HR}}_{\text{neut}} + U_{\theta}(\hat{\mathcal{T}}^t),
\end{equation}
where $\hat{\mathcal{T}}^t = D_{\text{tex}}(\mathbf{c}^t)$ is the Stage 1 prediction. This formulation focuses the network exclusively on expression-dependent, high-frequency deviations from the neutral state, substantially simplifying the learning problem and improving reconstruction of fine-grained detail. The network is supervised with a masked L1 loss in UV space:
\begin{equation}
    \mathcal{L}_{\text{up}} = \|\mathbf{M}^{\text{HR}} \odot (\mathcal{T}^t_{\text{HR}} - \hat{\mathcal{T}}^t_{\text{HR}})\|_1,
\end{equation}
where $\mathbf{M}^{\text{HR}}$ is the topology mask upsampled to 4K resolution.

The backbone of $U_{\theta}$ is a RepConv architecture \cite{Gankhuyag_2023_CVPR}, which employs multi-branch convolutions during training — combining $3{\times}3$, $1{\times}1$, and identity paths in parallel — to maximize representational capacity. At inference, all branches are fused into a single convolution via structural reparameterization~\cite{bhardwaj2022collapsible}, recovering the speed of a plain network with no architectural overhead. This makes $U_{\theta}$ both highly expressive for subject-specific detail recovery and efficient enough for real-time performance.

\subsection{Real-Time Monocular Tracking}
\label{ssec:monocular_tracking}
At inference, the pretrained geometry encoder of our tracker $E_\text{geom}$ regresses PCA coefficients $\mathbf{c}^t$ directly from a single lightstage camera frame, which are then decoded into a dynamic texture $\hat{\mathcal{T}}^t$ via $D_\text{tex}$ and upscaled to 4K by $U_\theta$ using our personalized upscaler. This unified approach delivers production-quality geometry and appearance in real time without any multi-camera capture or complex optimization.

\begin{figure}[!t]
    \centering
    \includegraphics[width=\linewidth]{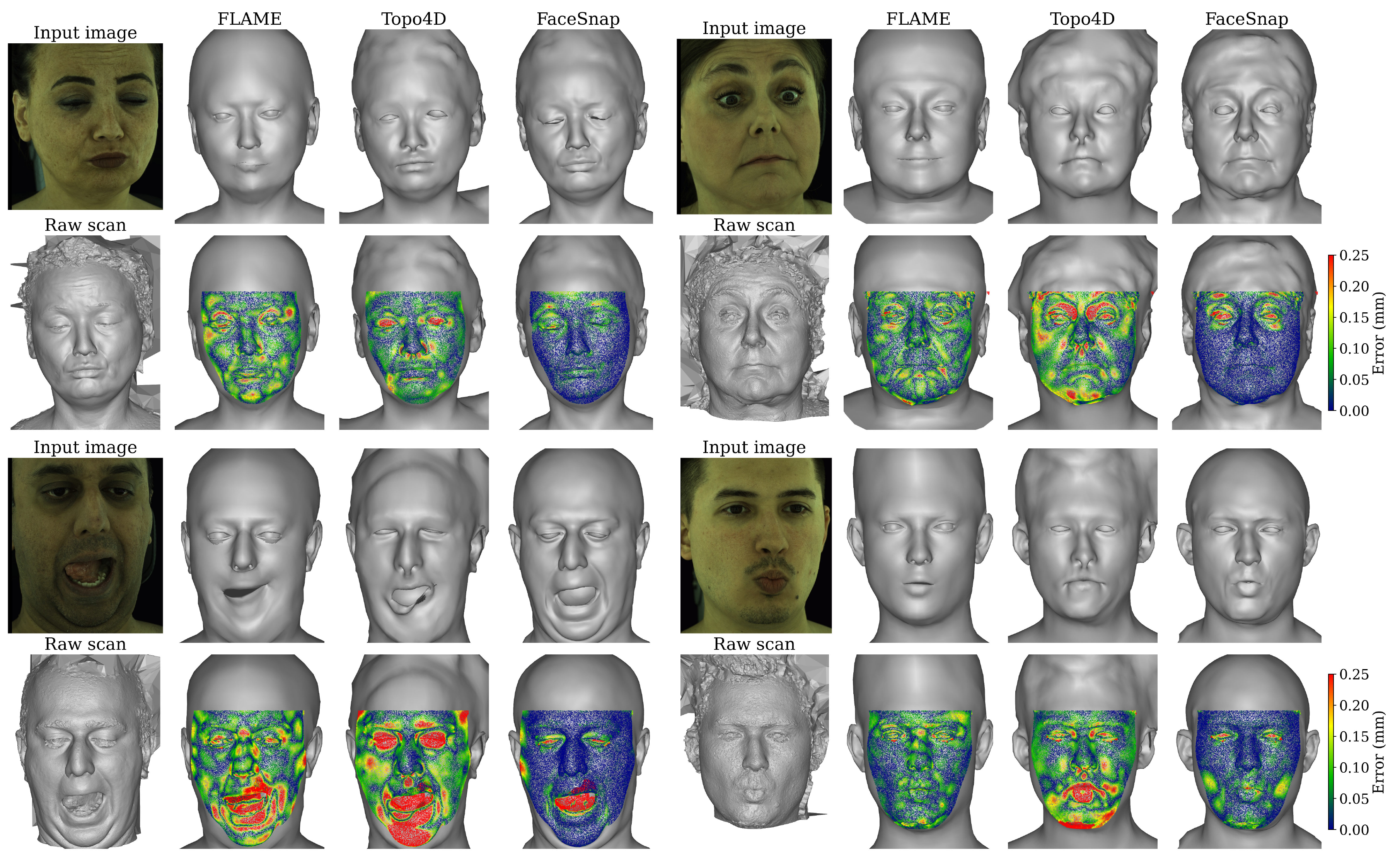}
    \caption{Visual comparison of mesh reconstruction quality across different subjects on the Multi4D benchmark.}
    \label{fig:geometric_comparison1}
\end{figure}

\definecolor{ourfull}{RGB}{214,229,255}
\definecolor{ourcoarse}{RGB}{173,202,255}

\begin{table}[!t]
\centering
\scriptsize
\setlength{\tabcolsep}{2.5pt}
\renewcommand{\arraystretch}{0.95}
\caption{Geometry comparison of multi-view face optimization methods using point-to-surface distance (mm, lower is better), $\pm$ per-frame standard deviation over each sequence.}
\vspace{-6pt}
\label{tab:geometry}
\begin{threeparttable}
\resizebox{\columnwidth}{!}{%
\begin{tabular}{@{}lrrrrrrrr@{\hspace{2pt}}}
\toprule
Method & \multicolumn{1}{c}{S1} & \multicolumn{1}{c}{S2} & \multicolumn{1}{c}{S3} & \multicolumn{1}{c}{S4} & \multicolumn{1}{c}{S5} & \multicolumn{1}{c}{S6} & \multicolumn{1}{c}{Overall} \\
\midrule
FLAME Fitting
 & \gvalo{0.643}\,{\tiny\textcolor{stdgray}{$\pm$0.151}}
 & \gvalo{1.35}\,{\tiny\textcolor{stdgray}{$\pm$0.237}}
 & \gvalo{0.625}\,{\tiny\textcolor{stdgray}{$\pm$0.194}}
 & \gvalo{0.600}\,{\tiny\textcolor{stdgray}{$\pm$0.207}}
 & \gvalo{0.672}\,{\tiny\textcolor{stdgray}{$\pm$0.0965}}
 & \gvalo{0.683}\,{\tiny\textcolor{stdgray}{$\pm$0.182}}
 & \gvalo{0.793} \\
Topo4D
 & \gvalo{1.28}\,{\tiny\textcolor{stdgray}{$\pm$0.465}}
 & \gvalo{1.71}\,{\tiny\textcolor{stdgray}{$\pm$0.780}}
 & \gvalo{1.21}\,{\tiny\textcolor{stdgray}{$\pm$0.382}}
 & \gvalo{1.07}\,{\tiny\textcolor{stdgray}{$\pm$0.504}}
 & \gvalo{0.641}\,{\tiny\textcolor{stdgray}{$\pm$0.175}}
 & \gvalo{1.21}\,{\tiny\textcolor{stdgray}{$\pm$0.483}}
 & \gvalo{1.24} \\
\rowcolor{ourcoarse} \textbf{Ours} (Base Fitting)
 & \gvalo{0.814}\,{\tiny\textcolor{stdgray}{$\pm$0.436}}
 & \gvalo{0.844}\,{\tiny\textcolor{stdgray}{$\pm$0.450}}
 & \gvalo{0.796}\,{\tiny\textcolor{stdgray}{$\pm$0.412}}
 & \gvalo{0.711}\,{\tiny\textcolor{stdgray}{$\pm$0.331}}
 & \gvalo{0.614}\,{\tiny\textcolor{stdgray}{$\pm$0.155}}
 & \gvalo{0.817}\,{\tiny\textcolor{stdgray}{$\pm$0.357}}
 & \gvalo{0.779} \\
\rowcolor{ourfull} \textbf{Ours} (Full)
 & \gvalbo{0.584}\,{\tiny\textcolor{stdgray}{$\pm$0.394}}
 & \gvalbo{0.505}\,{\tiny\textcolor{stdgray}{$\pm$0.404}}
 & \gvalbo{0.477}\,{\tiny\textcolor{stdgray}{$\pm$0.364}}
 & \gvalbo{0.420}\,{\tiny\textcolor{stdgray}{$\pm$0.278}}
 & \gvalbo{0.240}\,{\tiny\textcolor{stdgray}{$\pm$0.125}}
 & \gvalbo{0.508}\,{\tiny\textcolor{stdgray}{$\pm$0.345}}
 & \gvalbo{0.471} \\
\bottomrule
\end{tabular}%
}
\end{threeparttable}
\vspace{-10pt}
\end{table}

\section{Datasets and Benchmark}

\noindent\textbf{Multi4D Benchmark.} We propose Multi4D, a new benchmark based on the Multiface dataset~\cite{wuu2022multiface} to enable topology invariant mesh geometry comparison. It contains 4D lightstage captures of 6 subjects with varying facial characteristics performing a partial range of motion sequence between $\sim$1150 to $\sim$1500 frames each ($\sim$7600 frames total). Since Multiface does not provide ground truth scans, we extract raw scans using photogrammetry software \cite{realityscan2025} for each frame to serve as ground truth. For evaluation, we sample 300K vertices from each raw scan and clip them to include only the frontal facial region for point-to-surface distance computation. We publicly release the evaluation source code and ground-truth raw scans for all subjects. More details in the supp. material.

\noindent\textbf{Custom Lightstage Dataset.} \label{lightstage}
We capture data using a lightstage setup~\cite{debevec2012light} with 24 time-synchronized, calibrated cameras at 60 FPS and 4K resolution. Our dataset includes 3 subjects performing: (1) Range of Motion (ROM) sequences ($\sim$5 minutes) following FACS principles for profile creation, and (2) expressive sequences with semantically distinct spoken sentences (happy, angry, articulated, ~20 seconds each) for tracking evaluation. 
\section{Evaluation and Discussion}

%
\begin{figure}[!t]
    \centering
    \includegraphics[width=\linewidth]{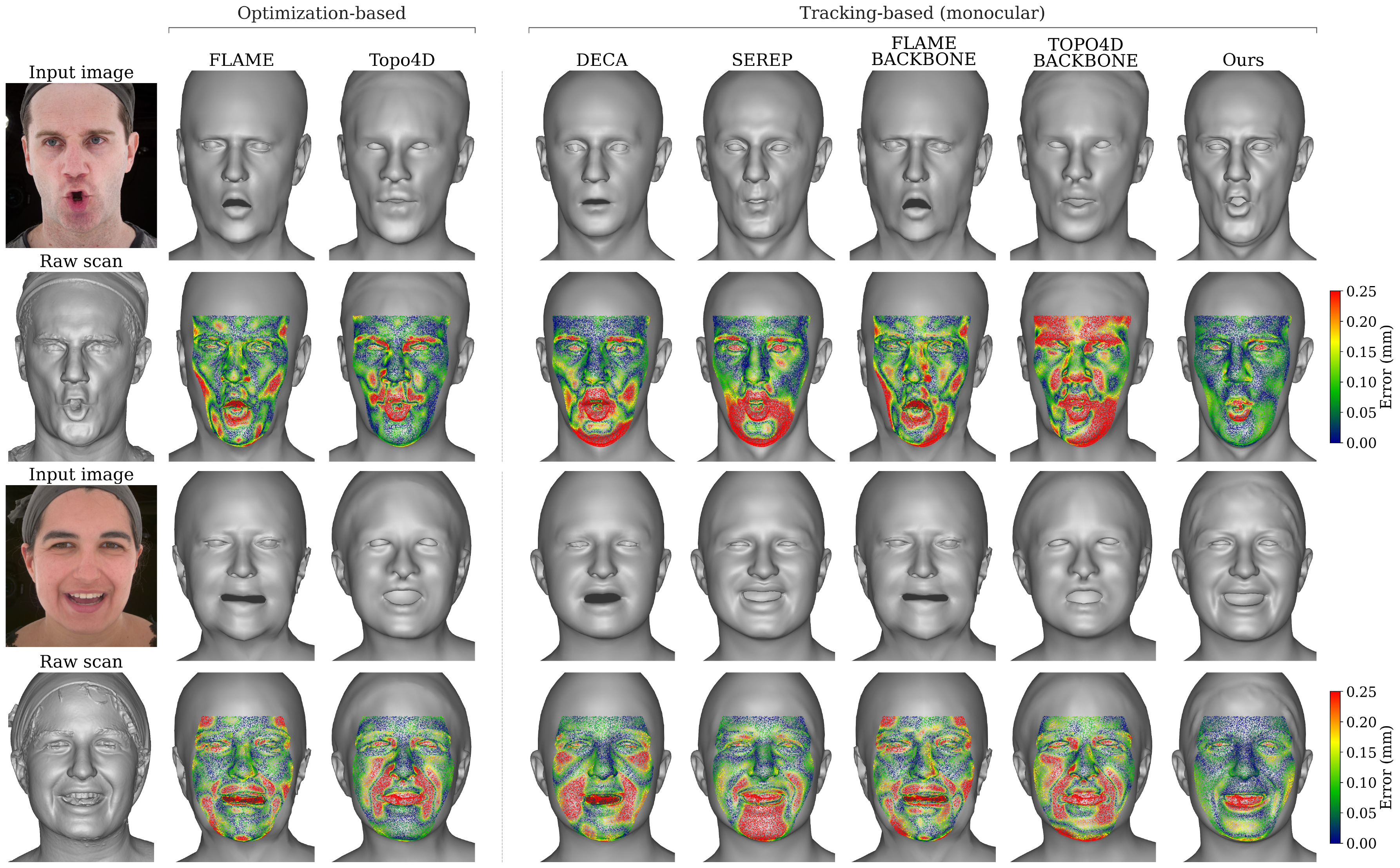}
    \caption{Qualitative geometric comparison for different methods.}
    \label{fig:geometric_comparison2}
\end{figure}

\begin{figure*}
    \centering
    \includegraphics[width=0.9\textwidth]{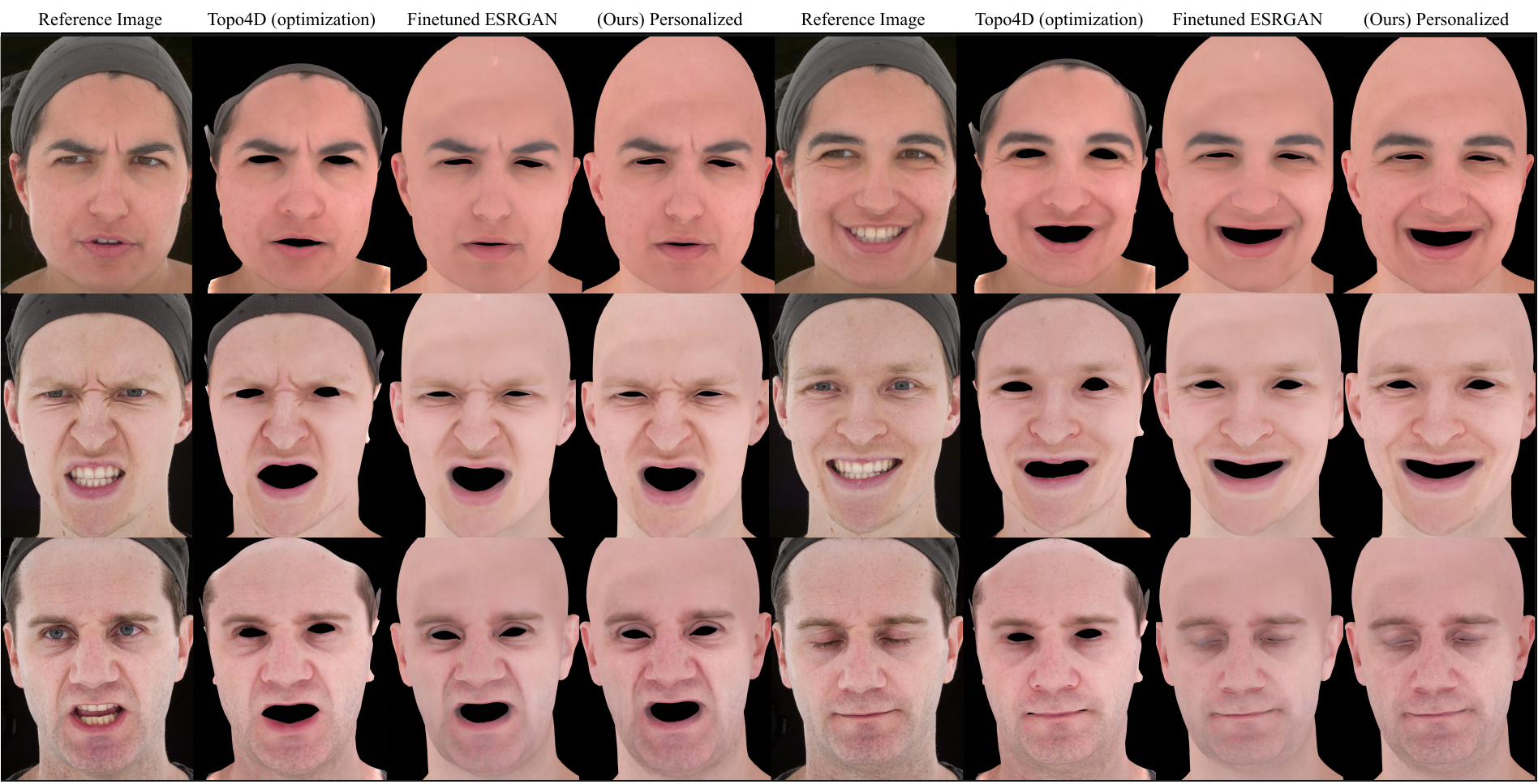}
    \vspace{-2pt}
  \caption{Qualitative comparison of appearance across subjects for FaceSnap and baseline methods on the custom lightstage dataset.}
    \label{fig:texture_comparison}
    \vspace{-2pt}
\end{figure*}

\subsection{Evaluation of our Multi-Stage Optimization}
\label{ssec:exp1}

We evaluate geometric accuracy of our optimization
(Sec.~\ref{ssec:optim}) against FLAME fitting~\cite{li2017learning},
which fits a FLAME model directly to raw photogrammetry scans, and Topo4D~\cite{li2024topo4d}, a Gaussian splatting-based per-frame optimization requiring a manually registered neutral mesh, part-wise facial annotations, and 8K texture and multi-view Lightstage images as input. Both share a comparable initialization burden to ours. We report point-to-surface distance against ground-truth photogrammetry scans on Multi4D in \autoref{tab:geometry}.

Our optimization achieves the best overall accuracy (0.47mm), improving 40.7\% over FLAME fitting (0.79mm) and 62.2\% over Topo4D (1.24mm). Topo4D shows high mean error and variance across subjects (range: 0.64--1.71mm), indicating sensitivity to rapid expression dynamics in Multiface sequences. FLAME is constrained by parametric model capacity, with reduced accuracy in perceptually critical region; lips, eyes, jaw, and chin, as seen in  \autoref{fig:geometric_comparison1}. The optimization first fits a parametric expression model (base fitting), then refines per-vertex displacements to recover fine-scale detail (vertex refinement). The gap between base fitting and full (0.78mm$\rightarrow$0.47mm) confirms that vertex refinement meaningfully improves geometric fidelity beyond what the parametric model alone can capture. Ablations of notable loss components, including the lip contour loss, are provided in the supplementary material.

\subsection{Evaluation of Real-Time Tracker}
\label{ssec:exp2}
\begin{table}[!t]
\centering
\scriptsize
\setlength{\tabcolsep}{2pt}
\renewcommand{\arraystretch}{0.95}
\caption{Geometric comparison of face tracking methods using average point-to-surface distance (mm, lower is better), $\pm$ average per-frame standard deviation over each sequence.}
\vspace{-6pt}
\label{tab:geometry2}
\begin{threeparttable}
\resizebox{\columnwidth}{!}{%
\begin{tabular}{@{}lrrrrrrrrrr@{\hspace{2pt}}}
\toprule
 & \multicolumn{3}{c}{Subject 1} & \multicolumn{3}{c}{Subject 2} & \multicolumn{3}{c}{Subject 3} & \\
\cmidrule(lr){2-4}\cmidrule(lr){5-7}\cmidrule(lr){8-10}
Method & \multicolumn{1}{c}{Artic.} & \multicolumn{1}{c}{Angry} & \multicolumn{1}{c}{Happy} & \multicolumn{1}{c}{Artic.} & \multicolumn{1
}{c}{Angry} & \multicolumn{1}{c}{Happy} & \multicolumn{1}{c}{Artic.} & \multicolumn{1}{c}{Angry} & \multicolumn{1}{c}{Happy} & \multicolumn{1}{c}{Overall} \\
\midrule
\rowcolor{gray!15} \multicolumn{11}{@{}l}{\hspace{2pt}\textit{Per-frame,Multi-View Optimization }} \\
FLAME Fitting
 & \gvalb{0.691}\,{\tiny\textcolor{stdgray}{$\pm$0.124}}
 & \gvalb{0.685}\,{\tiny\textcolor{stdgray}{$\pm$0.059}}
 & \gvalb{0.721}\,{\tiny\textcolor{stdgray}{$\pm$0.046}}
 & \gval{1.08}\,{\tiny\textcolor{stdgray}{$\pm$0.172}}
 & \gval{1.05}\,{\tiny\textcolor{stdgray}{$\pm$0.047}}
 & \gval{1.07}\,{\tiny\textcolor{stdgray}{$\pm$0.043}}
 & \gval{0.982}\,{\tiny\textcolor{stdgray}{$\pm$0.078}}
 & \gval{0.978}\,{\tiny\textcolor{stdgray}{$\pm$0.073}}
 & \gval{0.976}\,{\tiny\textcolor{stdgray}{$\pm$0.036}}
 & \gvalb{0.910}\,{\tiny\textcolor{stdgray}{$\pm$0.174}} \\
Topo4D
 & \gval{0.899}\,{\tiny\textcolor{stdgray}{$\pm$0.123}}
 & \gval{0.908}\,{\tiny\textcolor{stdgray}{$\pm$0.085}}
 & \gval{0.854}\,{\tiny\textcolor{stdgray}{$\pm$0.083}}
 & \gvalb{1.09}\,{\tiny\textcolor{stdgray}{$\pm$0.152}}
 & \gvalb{1.02}\,{\tiny\textcolor{stdgray}{$\pm$0.096}}
 & \gvalb{1.02}\,{\tiny\textcolor{stdgray}{$\pm$0.129}}
 & \gvalb{0.903}\,{\tiny\textcolor{stdgray}{$\pm$0.169}}
 & \gvalb{0.861}\,{\tiny\textcolor{stdgray}{$\pm$0.122}}
 & \gvalb{0.822}\,{\tiny\textcolor{stdgray}{$\pm$0.071}}
 & \gval{0.920}\,{\tiny\textcolor{stdgray}{$\pm$0.143}} \\
\midrule
\rowcolor{gray!15} \multicolumn{11}{@{}l}{\hspace{2pt}\textit{Feed-forward Methods}} \\
SEREP
 & \gval{1.78}\,{\tiny\textcolor{stdgray}{$\pm$0.269}}
 & \gval{1.91}\,{\tiny\textcolor{stdgray}{$\pm$0.209}}
 & \gval{2.10}\,{\tiny\textcolor{stdgray}{$\pm$0.265}}
 & \gval{1.53}\,{\tiny\textcolor{stdgray}{$\pm$0.304}}
 & \gval{1.40}\,{\tiny\textcolor{stdgray}{$\pm$0.311}}
 & \gval{1.30}\,{\tiny\textcolor{stdgray}{$\pm$0.289}}
 & \gval{1.16}\,{\tiny\textcolor{stdgray}{$\pm$0.294}}
 & \gval{1.43}\,{\tiny\textcolor{stdgray}{$\pm$0.317}}
 & \gval{1.65}\,{\tiny\textcolor{stdgray}{$\pm$0.316}}
 & \gval{1.59}\,{\tiny\textcolor{stdgray}{$\pm$0.409}} \\
DECA
 & \gval{3.06}\,{\tiny\textcolor{stdgray}{$\pm$1.42\phantom{0}}}
 & \gval{2.73}\,{\tiny\textcolor{stdgray}{$\pm$2.10\phantom{0}}}
 & \gval{3.33}\,{\tiny\textcolor{stdgray}{$\pm$1.44\phantom{0}}}
 & \gval{1.83}\,{\tiny\textcolor{stdgray}{$\pm$1.08\phantom{0}}}
 & \gval{2.23}\,{\tiny\textcolor{stdgray}{$\pm$5.22$\dagger$}}
 & \gval{2.30}\,{\tiny\textcolor{stdgray}{$\pm$5.48$\dagger$}}
 & \gval{1.58}\,{\tiny\textcolor{stdgray}{$\pm$0.747}}
 & \gval{2.12}\,{\tiny\textcolor{stdgray}{$\pm$1.01\phantom{0}}}
 & \gval{2.21}\,{\tiny\textcolor{stdgray}{$\pm$2.01\phantom{0}}}
 & \gval{2.38}\,{\tiny\textcolor{stdgray}{$\pm$2.82\phantom{0}}} \\
FLAME\_BACKBONE
 & \gval{1.11}\,{\tiny\textcolor{stdgray}{$\pm$0.231}}
 & \gval{2.22}\,{\tiny\textcolor{stdgray}{$\pm$0.732}}
 & \gval{2.44}\,{\tiny\textcolor{stdgray}{$\pm$0.634}}
 & \gval{1.29}\,{\tiny\textcolor{stdgray}{$\pm$0.113}}
 & \gval{1.45}\,{\tiny\textcolor{stdgray}{$\pm$0.308}}
 & \gval{1.46}\,{\tiny\textcolor{stdgray}{$\pm$0.196}}
 & \gval{1.26}\,{\tiny\textcolor{stdgray}{$\pm$0.166}}
 & \gval{1.74}\,{\tiny\textcolor{stdgray}{$\pm$0.402}}
 & \gval{2.40}\,{\tiny\textcolor{stdgray}{$\pm$0.549}}
 & \gval{1.75}\,{\tiny\textcolor{stdgray}{$\pm$0.663}} \\
TOPO4D\_BACKBONE
 & \gval{1.44}\,{\tiny\textcolor{stdgray}{$\pm$0.335}}
 & \gval{2.12}\,{\tiny\textcolor{stdgray}{$\pm$0.693}}
 & \gval{2.50}\,{\tiny\textcolor{stdgray}{$\pm$0.617}}
 & \gval{1.32}\,{\tiny\textcolor{stdgray}{$\pm$0.198}}
 & \gval{1.80}\,{\tiny\textcolor{stdgray}{$\pm$0.368}}
 & \gval{1.48}\,{\tiny\textcolor{stdgray}{$\pm$0.169}}
 & \gval{1.47}\,{\tiny\textcolor{stdgray}{$\pm$0.266}}
 & \gval{2.02}\,{\tiny\textcolor{stdgray}{$\pm$0.549}}
 & \gval{3.09}\,{\tiny\textcolor{stdgray}{$\pm$0.697}}
 & \gval{1.97}\,{\tiny\textcolor{stdgray}{$\pm$0.752}} \\
Ours (Tracking)
 & \gvalb{0.966}\,{\tiny\textcolor{stdgray}{$\pm$0.314}}
 & \gvalb{1.08}\,{\tiny\textcolor{stdgray}{$\pm$0.211}}
 & \gvalb{1.33}\,{\tiny\textcolor{stdgray}{$\pm$0.132}}
 & \gvalb{0.867}\,{\tiny\textcolor{stdgray}{$\pm$0.175}}
 & \gvalb{0.754}\,{\tiny\textcolor{stdgray}{$\pm$0.129}}
 & \gvalb{0.897}\,{\tiny\textcolor{stdgray}{$\pm$0.195}}
 & \gvalb{0.687}\,{\tiny\textcolor{stdgray}{$\pm$0.189}}
 & \gvalb{0.786}\,{\tiny\textcolor{stdgray}{$\pm$0.269}}
 & \gvalb{0.842}\,{\tiny\textcolor{stdgray}{$\pm$0.311}}
 & \gvalb{0.923}\,{\tiny\textcolor{stdgray}{$\pm$0.294}} \\
\bottomrule
\end{tabular}%
}
\begin{tablenotes}[flushleft]
  \item {\fontsize{5pt}{6pt}\selectfont $\dagger$\,DECA std exceeds the mean on these sequences, reflecting frequent failure frames.}
\end{tablenotes}
\end{threeparttable}
\end{table}
In this section, we evaluate our real-time tracker on lightstage data. We build a personalized tracking model from the ROM sequence using the lightstage data described in \autoref{lightstage} (custom lightstage data), and assess it on 9 expressive sequences, along three axes.

\noindent\textbf{Comparison to monocular trackers} We compare our monocular real-time tracker against state-of-the-art monocular tracking methods, DECA and SEREP, which are inference-based models trained on large-scale data.Both the methods are provided the subject neutral. For more details please see the supp mat.

\noindent\textbf{Reference comparison to multi-view optimization} As an upper-bound reference, we also compare against FLAME fitting and Topo4D, which recover geometry through per-frame optimization over  full camera array. Unlike our tracker, which infers geometry from a single camera at test time, these methods use all views and optimize each frame independently. We include them to show that our single-camera inference remains on par with full multi-view optimization, despite operating monocularly.

\noindent\textbf{Effect of the backbone} Our personalized tracking model is built on the registered meshes produced by our one-time optimization (\autoref{ssec:optim}). To isolate the contribution of this step, we replace it with alternative registration methods while keeping the downstream pipeline fixed. Specifically, we use FLAME fitting to register the ROM sequence, then build a personalized tracking model from the resulting meshes following \autoref{ssec:subject_profile}; we apply the same procedure with Topo4D. We refer to these variants as \texttt{FLAME\_BACKBONE} and \texttt{TOPO4D\_BACKBONE}. Since only the registration backbone differs, this comparison directly measures its effect on final tracker performance (details in supp. mat).

\noindent\textbf{Results.} FaceSnap ($0.92$mm) outperforms all feed-forward methods by a substantial margin. Notably, it surpasses SEREP ($1.59$mm) even though both receive identical neutral input and production-quality training data, confirming that subject-specific lightstage capture cannot be replaced by generic feed-forward methods. The backbone variants \texttt{FLAME\_BACKBONE} ($1.75$mm) and \texttt{TOPO4D\_BACKBONE} ($1.97$mm) perform worse than FaceSnap despite sharing its architecture, showing that tracker quality is driven by the quality of the registered meshes used to train the personalized model, not by architecture alone. Finally, FaceSnap is on par with the multi-view per-frame optimization baselines (FLAME: $0.91$mm, Topo4D: $0.92$mm), while using only a single-camera video sequence. However, as shown in \autoref{fig:geometric_comparison2}, mean point-to-surface scores alone do not capture localized errors in
perceptually critical regions where FaceSnap preserves lip shape and
identity detail better.

\definecolor{ourrow}{RGB}{230,230,230}

\begin{table}[!t]
\centering
\scriptsize
\setlength{\tabcolsep}{3.5pt}
\renewcommand{\arraystretch}{0.88}
\caption{Appearance quality comparison across subjects and expressions,
aggregated over 3 camera views. Shaded row = our method.
$\uparrow$ higher is better, $\downarrow$ lower is better.
$^*$Best perceptual similarity (LPIPS).}
\label{tab:appearance}

\vspace{-5pt}
\resizebox{\columnwidth}{!}{%
\begin{tabular}{llcccccccccc}
\toprule
 & & \multicolumn{3}{c}{Subject 1} & \multicolumn{3}{c}{Subject 2} & \multicolumn{3}{c}{Subject 3} & \\
\cmidrule(lr){3-5}\cmidrule(lr){6-8}\cmidrule(lr){9-11}
Metric & Method & Art. & Ang. & Hap. & Art. & Ang. & Hap. & Art. & Ang. & Hap. & Overall \\
\midrule
\multirow{3}{*}{SSIM $\uparrow$}
 & Topo4D & 0.955 & 0.952 & 0.952 & 0.978 & 0.979 & 0.978 & 0.943 & 0.940 & 0.939 & 0.957 \\
 & Upscaler (Face-Tuned) & \textbf{0.963} & \textbf{0.962} & \textbf{0.959} & \textbf{0.980} & \textbf{0.982} & \textbf{0.980} & \textbf{0.952} & \textbf{0.947} & \textbf{0.944} & \textbf{0.963} \\
\rowcolor{ourrow} & Ours (Personalised) & 0.960 & 0.958 & 0.955 & 0.979 & 0.980 & 0.979 & 0.946 & 0.941 & 0.939 & 0.960 \\
\midrule
\multirow{3}{*}{PSNR $\uparrow$}
 & Topo4D & \textbf{39.8} & \textbf{39.8} & \textbf{39.6} & \textbf{40.6} & \textbf{41.1} & \textbf{41.2} & \textbf{38.9} & \textbf{38.3} & \textbf{38.5} & \textbf{39.8} \\
 & Upscaler (Face-Tuned) & 38.0 & 37.2 & 37.0 & 38.9 & 38.9 & 39.0 & 37.1 & 36.5 & 36.6 & 37.7 \\
\rowcolor{ourrow} & Ours (Personalised) & 37.7 & 37.0 & 36.7 & 38.8 & 38.8 & 39.0 & 36.9 & 36.4 & 36.4 & 37.5 \\
\midrule
\multirow{3}{*}{LPIPS $\downarrow$}
 & Topo4D & 0.0543 & 0.0528 & 0.0579 & 0.0329 & 0.0300 & 0.0316 & 0.0627 & 0.0670 & 0.0632 & 0.0503 \\
 & Upscaler (Face-Tuned) & 0.0723 & 0.0758 & 0.0807 & 0.0538 & 0.0542 & 0.0531 & 0.0758 & 0.0807 & 0.0813 & 0.0698 \\
\rowcolor{ourrow} & Ours (Personalised)$^*$ & \textbf{0.0516} & \textbf{0.0560} & \textbf{0.0608} & \textbf{0.0312} & \textbf{0.0296} & \textbf{0.0310} & \textbf{0.0579} & \textbf{0.0620} & \textbf{0.0669} & \textbf{0.0497} \\
\bottomrule
\end{tabular}%
}
\vspace{-15pt}
\end{table}

\subsection{Evaluation of Personalized Residual Upscaler}
\label{ssec:exp3}
We evaluate the fidelity of the tracker's predicted 4K texture (see \autoref{upscaler}) via novel-view synthesis. For each subject (3 identities, 9 expressive sentences in total), we render the ground-truth photogrammetry scans from three novel viewpoints at 4K resolution and use these as reference. We compare three configurations spanning a range of capture and supervision costs: (i) Topo4D, which performs per-frame photometric optimization over the full multi-view array; (ii) our monocular tracker with ESRGAN \cite{wang2021real}, a state-of-the-art upscaler fine-tuned on a large-scale lightstage dataset of 1{,}000 subjects; and (iii) our monocular tracker with our personalized upscaler, trained solely on the target subject's ROM data.

Despite operating from a single monocular view, our personalized upscaler matches Topo4D's full multi-view capture in perceptual quality, achieving comparable LPIPS (0.0497 vs.\ 0.0503). SSIM is near-saturated across all methods ($<$0.01 spread), confirming that it does not discriminate at this fidelity. FaceSnap's lower PSNR relative to Topo4D (37.52 vs.\ 39.75 ,dB) is consistent with the perception-distortion tradeoff~\cite{blau2018perception}: methods optimizing for perceptual quality necessarily sacrifice some pixel-level fidelity. The contrast is decisive against the generic upscaler: our personalized model improves LPIPS by a substantial margin (0.0497 vs.\ 0.0698 for ESRGAN, fine-tuned on 1{,}000 subjects), demonstrating that single-subject ROM training recovers high-frequency identity detail that large-scale generic training cannot, as shown in \autoref{fig:texture_comparison}.

\subsection{Real-Time Performance and Efficiency}
\label{ssec:exp4}
FaceSnap's real-time monocular tracker predicts full geometry and 4K texture in \textbf{12\,ms per frame} on an NVIDIA RTX A6000, comprising geometry inference (0.6\,ms), appearance decoding (2.9\,ms), and 4K upscaling (8.5\,ms). Measured as end-to-end time to produce a tracked frame, this is a \textbf{5{,}000$\times$
 speedup} over Topo4D ($\sim60$ s) and a \textbf{25{,}000
$\times$ speedup} over the MVS+ICP registration used in production pipelines~\cite{faceform2025} (300--600\,s depending on expression complexity). Crucially, these gains come from a single camera rather than a full multi-view array: beyond runtime, FaceSnap removes per-session multi-camera capture and photogrammetry processing entirely.
\vspace{-5px}
\section{Conclusion}
\label{sec:conclusion}

We presented FaceSnap, an end-to-end framework that amortizes the cost
of high-fidelity lightstage facial capture into a reusable personalized subject model. Our one-time multi-stage optimization achieves superior geometric accuracy through non-linear expression modeling, vertex refinement, and a novel lip contour loss. The personalized tracker matches full multi-view optimization baselines at 83 fps from a single camera, with subject-specific appearance recovering high-frequency detail that generic models struggle to reproduce. We additionally introduce Multi4D, a public benchmark enabling topology-invariant evaluation of 4D facial reconstruction methods. Together, these contributions reduce the infrastructure and iteration cost of production facial capture while providing standardized tools for future research. Limitations and future work are discussed in the supplementary material.
\section{Acknowledgments}
This work was made possible through the financial support of the Natural Sciences and Engineering Research Council of Canada (NSERC) and MITACS, awarded through the Alliance-Mitacs program under Grant No. ALLRP 589317-23. We thank the teams behind Topo4D \cite{li2024topo4d}, FLAME Fitting \cite{li2017learning}, and DECA \cite{feng2021learning} for their open-source releases, and the authors of SEREP \cite{josi2024serep} for sharing their code with us. These resources enabled the comparisons and analyses reported in this work. We also thank the MultiFace team \cite{wuu2022multiface} for making their dataset available to build our benchmark upon.

{
    \small
    \bibliographystyle{ieeenat_fullname}
    \bibliography{main}
}
\newpage
\appendix
\setcounter{section}{0}
\setcounter{figure}{0}
\setcounter{table}{0}
\setcounter{equation}{0}

\section*{Supplementary Material}
\section{Ablation Studies}
\subsection{Lip Countour Loss ablation}
\begin{figure*}
    \centering
    \includegraphics[width=\textwidth]{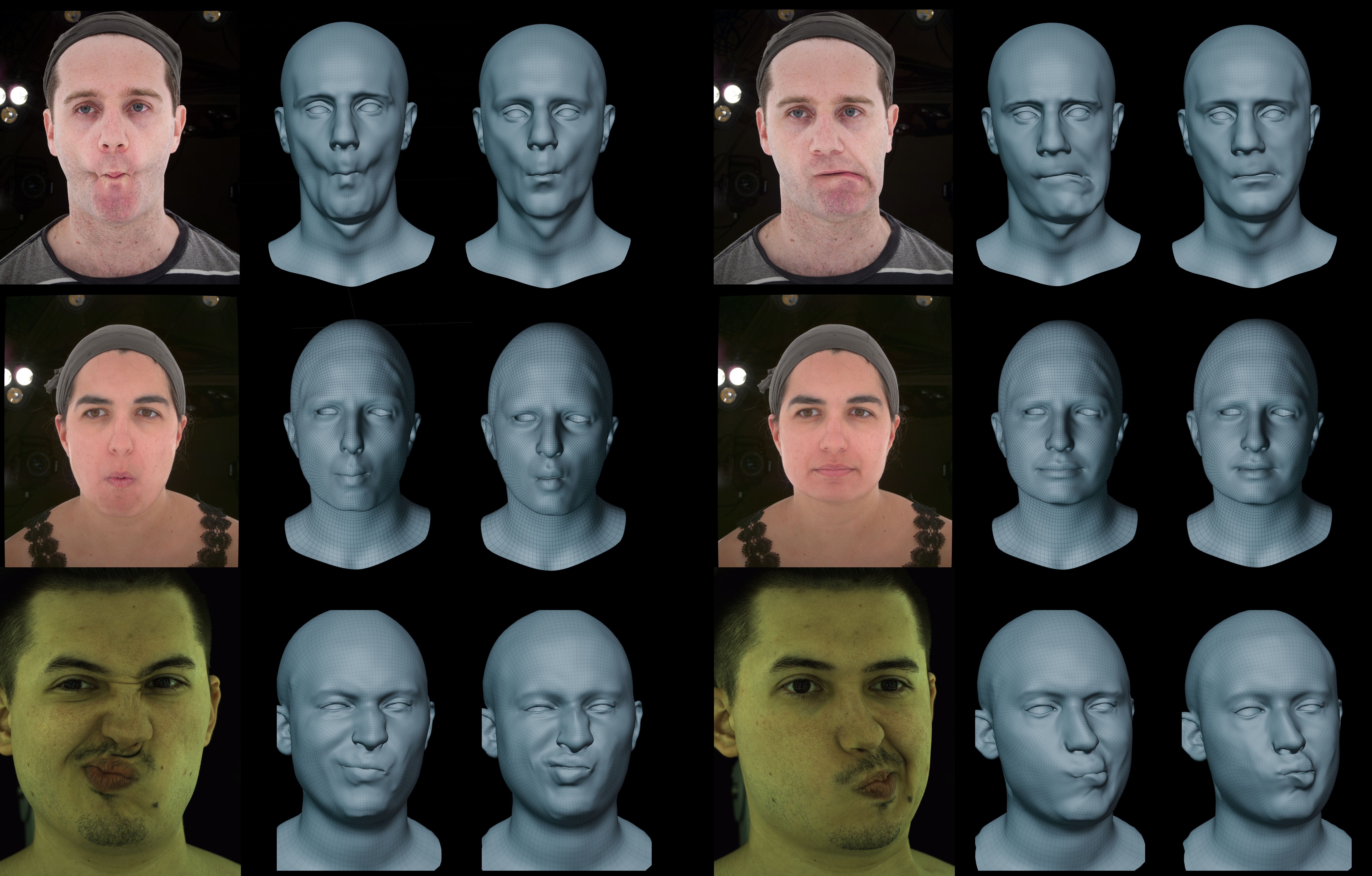}
    \caption{Ablation study demonstrating the importance of the lip contour loss for accurate lip tracking. From left to right: Input image, result with lip contour loss, result without lip contour loss. Note the improved lip closure and contour preservation when using the lip contour loss.}
    \label{fig:ablation_lip}
\end{figure*}

To demonstrate the effectiveness of our proposed lip contour loss, we conduct an ablation study by removing $\mathcal{L}_{\text{lip}}$ (eq. 5 in the paper)  from eq. 6 and replacing it with a traditional sparse landmark loss using 51 fixed facial keypoints (we exclude the first 17 landmarks, corresponding to the jaw and chin contour as these boundary points are inherently ambiguous.) 
As shown in \autoref{fig:ablation_lip}, our contour-based approach significantly outperforms the landmark-based baseline in capturing subtle lip movements and maintain accurate lip contact. Unlike landmark-based methods that require fixed point correspondences, our approach avoids the correspondence problem entirely, enabling more flexible and accurate lip deformation, critical for high-fidelity facial animation.
\subsection{Regularization Loss Ablation}
\begin{figure*}
    \centering
    \includegraphics[width=\textwidth]{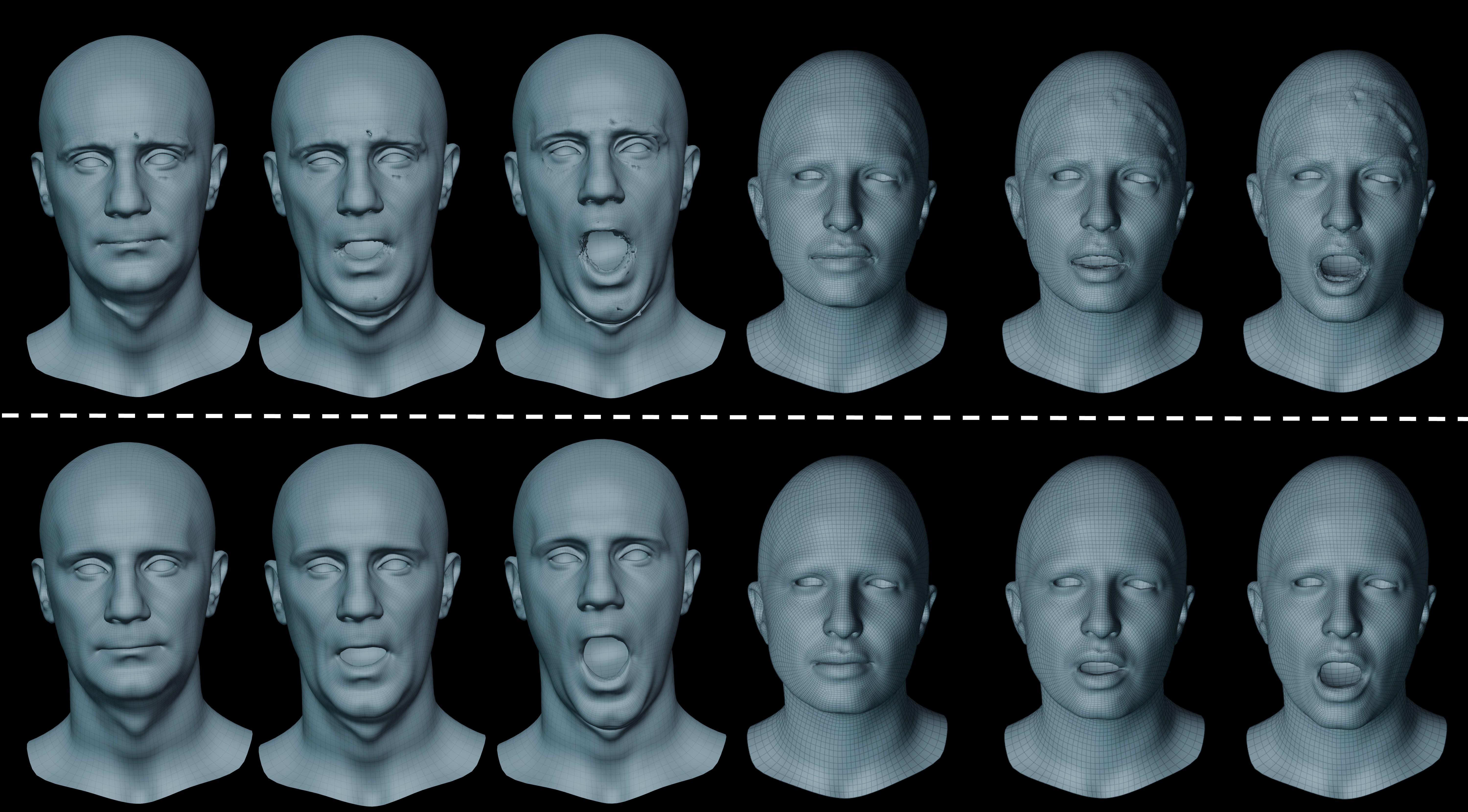}
    \caption{Ablation study demonstrating the importance of regularization losses. Top row: tracking results without regularization losses showing severe mesh degradation artifacts including inconsistent edge flows and surface discontinuities. Bottom row: tracking with full regularization showing stable mesh topology and smooth deformations.}
    \label{fig:ablation_reg}
\end{figure*}

To validate the importance of our regularization terms, we conduct an ablation study by removing all regularization losses ($\mathcal{L}_{\text{edge}}$, $\mathcal{L}_\text{lap}$, and $\mathcal{L}_\text{nc}$) from the overall objective function (Eq. 6 in the paper). As demonstrated in \autoref{fig:ablation_reg}, the absence of these regularization terms leads to severe mesh degradation artifacts. Without the part-based edge preservation loss $\mathcal{L}_{\text{edge}}$, we observe prominent sliding effects that produce inconsistent edge flows, particularly noticeable around critical facial regions (The edge flows are visualized as wireframe overlays on top of the geometry in \autoref{fig:ablation_reg}).
The removal of the normal consistency loss $\mathcal{L}_\text{nc}$ results in flipped face normals that create unrealistic surface orientations. Additionally, without the Laplacian loss $\mathcal{L}_\text{lap}$, the mesh exhibits jagged surface artifacts and unnatural discontinuities that violate smooth deformation assumptions. These results clearly demonstrate that our regularization losses are essential for maintaining mesh topology consistency and ensuring physically plausible facial reconstructions during optimization.

\section{Multi-Stage Optimization Details}

Our multi-stage optimization takes multi-view lightstage images and raw photogrammetry scans as input
and produces a sequence of registered meshes in a canonical topology, serving as pseudo-GT for the personalized subject model. The pipeline comprises three sequential stages; global registration, base fitting, and vertex refinement, each building on the previous. Here, we provide additional details for each stage.

\subsubsection{Global Registration}
The goal of global registration is to bring the raw photogrammetry
sequence $\{\mathcal{R}^t\}$ into the canonical coordinate space defined
by the subject neutral mesh $\mathcal{M}_{\mathrm{neutral}}$, so that
subsequent per-frame optimization stages operate in a consistent space.
We estimate a similarity transform $(\mathbf{R}_{\mathrm{g}},
\mathbf{T}_{\mathrm{g}}, s_{\mathrm{g}})$ via Procrustes
alignment~\cite{dryden2016statistical} on the neutral frame $t=0$.

Landmarks on $\mathcal{R}^0$ are obtained by rendering the raw scan
with its appearance texture into a 2D image, applying a 2D landmark
detector~\cite{Zhou_2023_CVPR}, and back-projecting detections to 3D
by intersecting rays with $\mathcal{R}^0$. Following~\cite{li2017learning},
we exclude the first 17 landmarks corresponding to the jaw and chin
contour, as these boundary points are inherently ambiguous due to
self-occlusion and soft tissue deformation. The estimated transform is
then applied to the full sequence $\{\mathcal{R}^t\}$, producing the
canonically aligned sequence $\{\widetilde{\mathcal{R}}^t\}$ used as
the registration target for all subsequent stages.

\subsection{Regularizaton losses}

The following regularization losses are used in Eq. 6 of the paper. 

\noindent{\textbf{Part-Based Edge Preservation Loss. }}
To preserve the local structure of the mesh and prevent vertex sliding effects, we use a part-based edge preservation loss, as in \cite{Bolkart2023Tempeh}, that constraint the edge vectors of the optimized mesh $\mathcal{X}^t$ to the previous mesh $\mathcal{X}^{t-1}$: 
\begin{align}
  \mathcal{L}_{\text{edge}} = \sum_{e \in E} w_e \|\mathbf{e}^t - \mathbf{e}^{t-1}\|^2  
\end{align}
where $E$ represents the set of mesh edges, $\mathbf{e}^t$ and $\mathbf{e}^{t-1}$ are  edge vectors obtained from $\mathcal{X}^t$ and $\mathcal{X}^{t-1}$ respectively.  $w_e$ are spatially-varying weights that emphasize preservation in critical facial regions such eye region and lips. 

\noindent{\textbf{Laplacian Loss. }}
To promote smooth deformations and penalize irregular or jaggy surface areas, especially for  regions known to be challenging for photogrammetry reconstruction, such as eyebrows and eyelids, we incorporate a Laplacian regularization: 
\begin{equation}
  \mathcal{L}_\text{lap} =  \|\delta_i - \delta_i^\text{neutral}\|^2  
\end{equation}
where $\delta_i$
is the Laplacian coordinate for vertex $i$, 
and $\delta_i^{neutral}$ is the corresponding Laplacian coordinate in the neutral mesh $\mathcal{M}_\text{neutral}$. The Laplacian coordinates effectively encode the local differential properties of the mesh, and preserving these properties leads to natural deformations.
%
%

\noindent{\textbf{Normal Consistency Loss. }}
To encourage smooth surface reconstruction and consistent face orientations, we apply a normal consistency loss to the reconstructed mesh $\mathcal{X}^t$.
\begin{equation}
\mathcal{L}_\text{nc} = \sum_{(f_0, f_1) \in \mathcal{E}} \left( 1 - \cos(\mathbf{n}_0, \mathbf{n}_1) \right),    
\end{equation}

where $f_0$ and $f_1$ are adjacent triangular faces sharing an edge, $\mathbf{n}_0$ and $\mathbf{n}_1$ are their respective face normals and $\mathcal{E}$ is the set of all adjacent face pairs.
\section{Implementation Details}
Our FaceSnap framework is implemented in PyTorch with CUDA enabled GPU.

\subsection{Multi-stage Optimization} For the multi-stage optimization (Sec. 3.1 in the paper), we use gradient descent across the the different steps (base fitting and vertex refinement). 
For base fitting stage (refer to Eq. 6 in the paper), we perform $100$ gradient descent step, with learning rate equal to $0.01$, with the following regularization terms: $\lambda^\text{base}_1 = 1 $, $\lambda^\text{base}_2 = 1e^{-06}$, $\lambda^\text{base}_3 = 2$, $\lambda^\text{base}_4 = 10000$, $\lambda^\text{base}_5 = 2.0$. For the vertex refinement stage (refer to Eq. 6 in the paper), we perform $300$ gradient descent step, with learning rate equal to $0.0001$ with the following regularization losses: $\lambda^\text{vert}_1 = 1 $, $\lambda^\text{vert}_2 = 1e^{-06}$, $\lambda^\text{vert}_3 = 2$, $\lambda^\text{vert}_4 = 20000$, $\lambda^\text{vert}_5 = 2.0$. The threshold for the automatic keyframe detector is $\tau_f = 0.18$ cm. For the adaptive point-to-surface loss, we use $\tau_s = 0.2$ cm and $\sigma = 1$.

\subsection{Personalized tracking Model}

\textbf{Geometry Model:} For the geometry model (Sec. 3.2.1 in the paper), we use a lightweight ResNet18 encoder, trained for 100 epochs with Adam optimizer ($\beta_1=0.9, \beta_2=0.999$), with a learning rate equal to $1\times 10^{-4}$, and a batch size of 8.

\noindent\textbf{Dynamic Appearance Model:} The model consists of two stages; low resolution prediction and high resolution up-scaling. 

\noindent\textbf{Expression-Conditioned Texture Synthesis (Low Res.)} (see Sec. 3.2.2, stage 1, in the paper) consists of two main components: a feature projection network and a convolutional decoder.

\begin{table}[h]
\centering
\caption{Dense Feature Projection Layers}
\begin{tabular}{|c|c|c|c|}
\hline
\textbf{Layer} & \textbf{Input} & \textbf{Output} & \textbf{Activation} \\
\hline
Linear 1 & 256 & 512 & LeakyReLU \\
Linear 2 & 512 & 1024 & LeakyReLU \\
Linear 3 & 1024 & 2048 & None \\
\hline
\end{tabular}
\label{tab:dense_layers}
\end{table}

\begin{table}[h]
\centering
\caption{Convolutional Decoder Architecture (per branch)}
\resizebox{\columnwidth}{!}{%
\begin{tabular}{|c|c|c|c|}
\hline
\textbf{Stage} & \textbf{Operations} & \textbf{Resolution} & \textbf{Channels} \\
\hline
1 & ConvT + Conv & $8 \times 8 \rightarrow 16 \times 16$ & $32 \rightarrow 64 \rightarrow 128$ \\
2 & ConvT + 2×Conv & $16 \times 16 \rightarrow 32 \times 32$ & $128 \rightarrow 128 \rightarrow 64 \rightarrow 192$ \\
3 & ConvT + 2×Conv & $32 \times 32 \rightarrow 64 \times 64$ & $192 \rightarrow 192 \rightarrow 96 \rightarrow 128$ \\
4 & ConvT + 2×Conv & $64 \times 64 \rightarrow 128 \times 128$ & $128 \rightarrow 128 \rightarrow 64 \rightarrow 64$ \\
5 & ConvT + 2×Conv & $128 \times 128 \rightarrow 256 \times 256$ & $64 \rightarrow 64 \rightarrow 42 \rightarrow 32$ \\
6 & ConvT + 2×Conv & $256 \times 256 \rightarrow 512 \times 512$ & $32 \rightarrow 42 \rightarrow 21 \rightarrow 3$ \\
\hline
\end{tabular}%
}
\label{tab:conv_decoder}
\end{table}

First, the model (Table~\ref{tab:dense_layers}) projects 256-dimensional PCA coefficients through three dense layers into a 2048-dimensional embedding. Then a convolutional decoder, progressively upsample from $8 \times 8$ to $512 \times 512$ resolution (Table~\ref{tab:conv_decoder}). Each stage uses transposed convolution for upsampling followed by regular convolutions with BatchNorm and LeakyReLU activations. ConvT denotes ConvTranspose2d. All convolutional layers are initialized with weights from $\mathcal{N}(0, 0.0002)$ and biases uniformly distributed in $[0.001, 0.0015]$. BatchNorm weights are initialized with $\mathcal{N}(1, 0.02)$ and biases set to $0.01$. The model is trained using the Adam optimizer with a learning rate of $5 \times 10^{-4}$ and a batch size of 16.

\begin{table}[h]
\centering
\caption{High-Resolution Residual Upscaler Architecture} 
\resizebox{\columnwidth}{!}{%
\begin{tabular}{|c|c|c|c|}
\hline
\textbf{Stage} & \textbf{Module} & \textbf{Channels} & \textbf{Resolution} \\
\hline
Head       & RepBlock (ReLU)    & $3 \rightarrow 64$   & $512 \times 512$ \\
Backbone   & 4$\times$ RepBlock (ReLU) & $64 \rightarrow 64$  & $512 \times 512$ \\
Transition & RepBlock (linear)  & $64 \rightarrow 192$ & $512 \times 512$ \\
Upsample   & PixelShuffle ($\times 8$) & $192 \rightarrow 3$ & $512 \times 512 \rightarrow 4096 \times 4096$ \\
\hline
\end{tabular}%
}
\label{tab:repconv_upscaler}
\end{table}

\noindent\textbf{High-Resolution Personalized Residual Upscaler:} (see Sec. 3.3.2, stage 2 in the paper)
Rather than predicting the full $4096 \times 4096$ UV texture directly, the model predicts a
subject-specific residual offset that is added to a pre-computed neutral texture.
The network (Table~\ref{tab:repconv_upscaler}) takes as input a $512 \times 512$ low-resolution residual UV map; the difference between the current expression texture and the neutral texture and upscales it by a factor of $8\times$ using a re-parameterizable
convolutional backbone~\cite{bhardwaj2022collapsible} followed by PixelShuffle.
The backbone consists of a head block, four residual feature-refinement blocks, and a
transition block with a skip connection from the head output, i.e.\
$\mathbf{y} = \text{Transition}(\text{Backbone}(\mathbf{y}_0) + \mathbf{y}_0)$,
where $\mathbf{y}_0 = \text{Head}(\mathbf{x})$.
Each RepBlock \cite{Gankhuyag_2023_CVPR} uses a $3\!\times\!3$ convolution through a 256-channel bottleneck followed
by a $1\!\times\!1$ projection; at inference, both branches are folded into a single
$3\!\times\!3$ convolution via structural re-parameterization, incurring no additional
runtime cost.
The model is trained with a masked L1 loss (restricted to the facial foreground region)
using AdamW ($\text{lr}=5\times10^{-4}$, $\beta=(0.9,\,0.999)$, weight decay $10^{-4}$)
with cosine annealing and mixed-precision training.

\section{Benchmark Design}
\noindent\textbf{Producing the data:} We design the Multi4D benchmark to enable direct comparison in the topology space of each method without requiring retargeting, establishing a standardized protocol for geometry evaluation. Multi4D represents the first publicly available benchmark designed to evaluate 4D facial capture methods on range of motion sequences, facilitating fair comparisons between approaches with different underlying topologies. To obtain the raw meshes used as ground truth, we run a commercial photogrammetry tool \cite{realityscan2025} on 38 camera views from the Multiface dataset \cite{wuu2022multiface}, covering 360 degrees around the full head with most cameras focused on the front and side views (please refer to Section 5 in the paper). This comprehensive multi-view reconstruction provides high-fidelity reference geometry for evaluation.
\begin{figure}
    \centering
    \includegraphics[width=.5\textwidth]{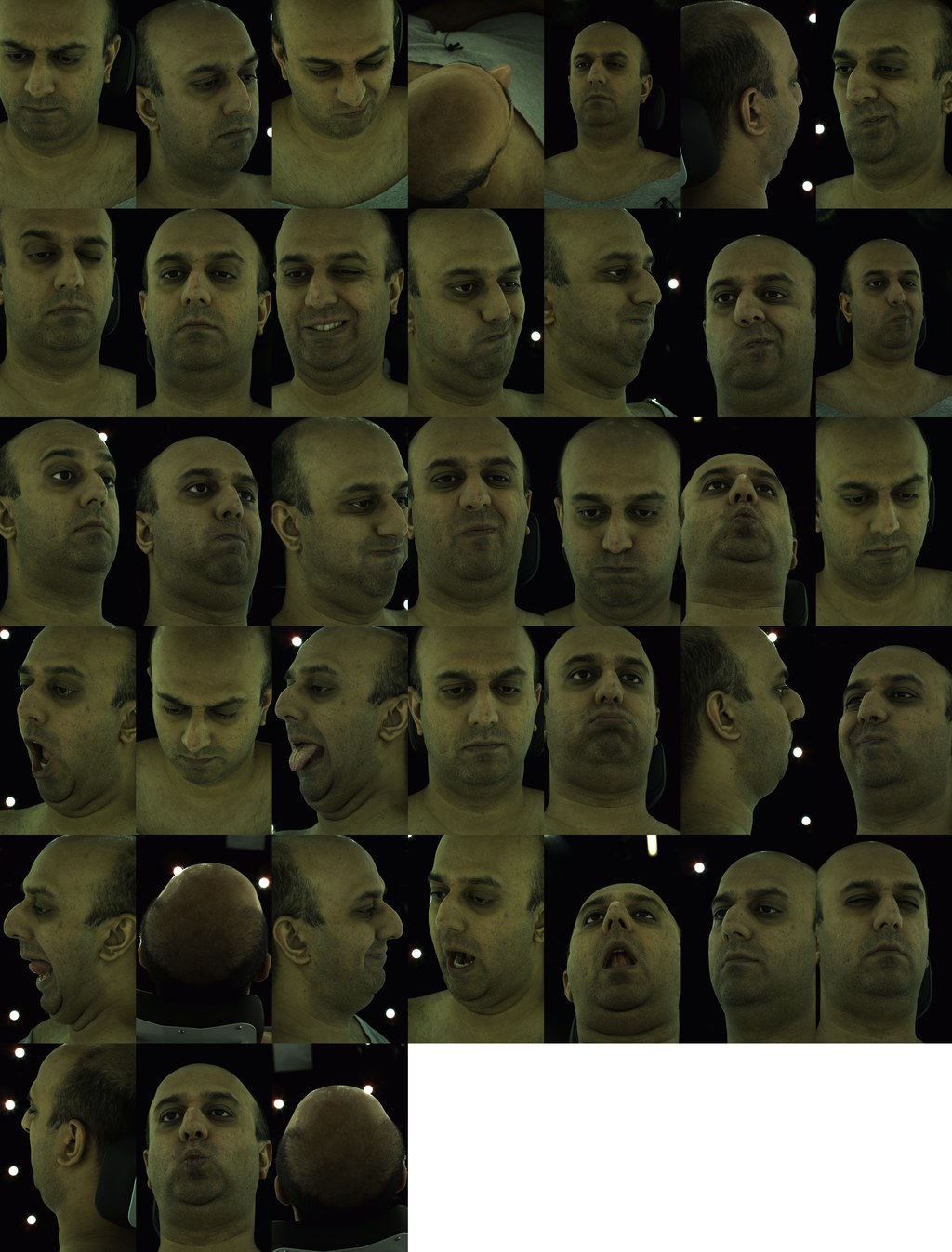}
    \caption{Multi4D benchmark camera setup: Sample images from the 38 cameras with different expressions used to run photogrammetry to obtain the raw scans used as ground truth for evaluation.}
    \label{fig:multi4d}
\end{figure}

\noindent\textbf{Metric calculation:} We evaluate every other frame for efficiency and sample 300{,}000 points uniformly from each raw scan. Points are clipped to a subject-specific axis-aligned bounding box covering the frontal facial region (forehead, eyes, ears, cheeks, and jaw). The bounding box is defined as six scalars $[x_\text{min}, x_\text{max}, y_\text{min}, y_\text{max}, z_\text{min}, z_\text{max}]$ in raw scan space and is set per subject; the exact values are provided in the released evaluation code. Point-to-surface distance is then computed between these clipped points and each method's reconstructed mesh. The full protocol described in \autoref{sec:eval_protocol} applies to both the Multi4D benchmark (Sec.~5.1) and the expressive sequence evaluation (Sec.~5.2).

We note  that for Topo4D~\cite{li2024topo4d} evaluation, we exclude cameras capturing back views since their method only optimizes for the frontal facial region. For the FLAME fitting method~\cite{li2017learning}, we employ an off-the-shelf landmark detector \cite{Zhou_2023_CVPR} to extract facial landmarks that are used by FLAME fitting process to rigidly aligning raw scans to their canonical coordinate system. Finally, we publicly release the evaluation source code and the ground-truth raw scans (obtained via photogrammetry) for each ROM sequence of the five subjects.

\section{Point-to-Surface Evaluation Protocol}
\label{sec:eval_protocol}

Both geometry evaluations (Sec.~5.1 and Sec.~5.2) follow the sampling and metric protocol described above. The key additional consideration for Sec.~5.2 is method alignment.

\noindent\textbf{Alignment to metric space.}
For per-frame multi-view optimization methods (FLAME Fitting, Topo4D, and our optimization), the known camera parameters bring each reconstructed mesh directly into raw scan space. For tracking-based methods (SEREP, DECA, \texttt{FLAME\_BACKBONE}, \texttt{TOPO4D\_BACKBONE}, and Ours), direct ICP to the raw scan is unreliable due to topological differences. Instead, each tracking method is aligned to its corresponding per-frame optimization result using point-to-point ICP (SVD-based, 200~iterations) on the first frame, then applied rigidly to all frames: SEREP, DECA, and Ours align to our optimization; \texttt{FLAME\_BACKBONE} aligns to FLAME Fitting; \texttt{TOPO4D\_BACKBONE} aligns to Topo4D.

%
\section{Limitations and future work}
Although FaceSnap achieves fast and efficient facial performance capture and outperforms state-of-the-art methods in both geometric accuracy and texture appearance quality, several aspects could benefit from further improvements. Our dynamic appearance model, optimized for real-time performance using a simple CNN architecture, operates at 512$\times$512 resolution and may lose fine details such as micro-wrinkles during the decoding process, which is a notable constraint compared to offline methods that can operate at higher resolutions. While the proposed personalized up-scaler recovers recovers most fine details in $4K$, regions near forehead are challenging. 

Additionally, the lip region presents particular challenges for our method, exhibiting subtle temporal artifacts due to occasional errors in photogrammetry or the automated face segmentation. Range of Motion (ROM) capture requires the subject to return to neutral state periodically to allow our automatic keyframing to detect neutral poses and reset the tracking to avoid drift. Finally, our evaluation is limited in subject diversity: optimization geometry is evaluated on Multi4D (6 subjects, Table 1, main paper), while monocular tracking and appearance are evaluated on our custom lightstage dataset (Table 2 and 3 using 3 subjects, 3 expressive sequences each, 9 sequences total). Broader validation across skin tone, age, and facial hair is needed to confirm generalization, particularly for the real-time tracking component; this is primarily constrained by the cost and limited availability of lightstage capture sessions rather than a methodological limitation.

As future work, we aim to incorporate facial intrinsic recovery to enable relighting by decomposing skin reflectance from light. Furthermore, enhancing our dynamic appearance model to produce higher resolution outputs with correctives to avoid temporal artifacts while maintaining the real-time performance aspect—which was a core design choice—represents an important direction for future development. Expanding evaluation to a larger and more demographically diverse subject pool is another important direction we leave for future work.

\section{Topo4D Initialization}
Topo4D~\cite{li2024topo4d} requires that the first frame of the sequence be perfectly aligned and deformed to match the raw scan before initiating their optimization process. To achieve this initialization, we use a commercial tool Wrap4D~\cite{faceform2025}, the same tool used by the original paper, to obtain a registered mesh and 8K texture for the first frame which is then used to start the optimization process. The wrapping process involves loading the raw scan and Topo4D template mesh, excluding certain polygons corresponding to eyes and mouth interior from registration. Wrap4D performs a two-stage alignment: first rigid alignment to transform the raw scan into the coordinate system of the template mesh, followed by non-rigid deformation to accurately fit the template to the raw scan geometry. Following the authors' recommendations, we scale the wrapped mesh to match Topo4D's template mesh scale and adjust camera calibration accordingly.

\end{document}

%% file: preamble.tex
\usepackage[dvipsnames]{xcolor}
